%% file: main.tex
\documentclass{article}
\usepackage{amssymb}
\usepackage{amsmath}
\usepackage{xspace}
\usepackage{graphicx}
\usepackage{wrapfig}
\PassOptionsToPackage{numbers,sort&compress,square}{natbib}

\usepackage{enumitem}  
\usepackage{float}
\usepackage{afterpage}
\usepackage{stfloats}
\usepackage{tabularx}
\usepackage{array}
\usepackage{makecell}
\usepackage{booktabs}
\usepackage{multirow}
\usepackage{dsfont}
\usepackage{hyperref}
\usepackage[dvipsnames]{xcolor}
\usepackage{xspace}
\usepackage{cite}
\usepackage{cuted}
\usepackage{caption}
\usepackage{bbm}

\usepackage{mdframed} 
\usepackage[T1]{fontenc} 
\usepackage{algorithm}
\usepackage{algpseudocode}
\usepackage{times}
\usepackage{threeparttable}
\usepackage{multirow}
\usepackage{multicol}
\usepackage{subcaption}
\usepackage{colortbl}
\usepackage{appendix}

\usepackage[final]{corl_2025} 

\title{Diffusion Dynamics Models with Generative State Estimation for Cloth Manipulation \vspace{-6pt}}

\author{
\textbf{Tongxuan Tian}$^{1*}$ \quad
\textbf{Haoyang Li}$^{1, 2*}$ \quad \\
\textbf{Bo Ai}$^{1}$ \quad 
\textbf{Xiaodi Yuan}$^{1}$ \quad 
\textbf{Zhiao Huang}$^{1, 2}$ \quad  
\textbf{Hao Su}$^{1, 2}$ \\ 
$^{*}$Equal contribution  \vspace{3pt}  \\
$^{1}$University of California San Diego, USA \quad $^{2}$Hillbot Inc, USA \vspace{2pt} \\
\href{https://uniclothdiff.github.io/}{\textbf{\textcolor{magenta}{https://uniclothdiff.github.io/}}} \vspace{-2pt}
} 
\input{text/marcos}

\begin{document}
\maketitle

\vspace{-0.5cm}
\input{text/000_abstract}

\input{text/010_introduction}
\input{text/020_related_work}
\input{text/030_method}
\input{text/040_experiments}
\input{text/050_conclusion}

\bibliography{main}
\clearpage
\input{text/060_appendices}

\end{document}

%% file: text/marcos.tex
\renewcommand{\eqref}[1]{Eqn~\ref{#1}}
\newcommand{\figref}[1]{Figure~\ref{#1}}

\newcommand{\tabref}[1]{Table~\ref{#1}}

\newcommand{\appendref}[1]{Appendix~\ref{#1}}

\newcommand{\eg}{\textrm{e.g.}}

\def\S{\mathcal{S}}
\def\A{\mathcal{A}}

\def\T{\mathcal{T}}
\def\T{\mathcal{J}}
\def\T{\mathcal{T}}

\def\J{\mathcal{J}}

\def\O{\mathcal{O}}

\DeclareMathOperator*{\argmin}{arg\,min}

\newcommand{\newtext}[1]{\textcolor{black}{#1}}

\def\ours{UniClothDiff\xspace}

\def\oursdynamics{DDM\xspace}
\def\oursperception{DPM\xspace}

\def\garmentnets{GarmentNets\xspace}
\def\medor{MEDOR\xspace}
\def\trtm{TRTM\xspace}
\def\dpmwoddpm{Transformer\xspace}
\def\ddmwoddpm{Transformer\xspace}

\def\gnn{GNN\xspace}
\def\simulator{Analytical Simulator\xspace}

\def\weblink{\urllink[pre = \bgroup\bf, post = \egroup]}

  {\begin{mdframed}[topline=false, bottomline=false, rightline=false,%
    linewidth=1pt,linecolor=black,%
    leftmargin=1cm,rightmargin=1cm,%
    innerleftmargin=10pt,innerrightmargin=10pt,%
    innertopmargin=0pt,innerbottommargin=0pt,%
    skipabove=\baselineskip,skipbelow=\baselineskip]%
   \fontfamily{ppl}\selectfont
  }%
  {\end{mdframed}}



%% file: text/000_abstract.tex
\begin{abstract}
Cloth manipulation is challenging due to its highly complex dynamics, near-infinite degrees of freedom, and frequent self-occlusions, which complicate both state estimation and dynamics modeling. 
Inspired by recent advances in generative models, we hypothesize that these expressive models can effectively capture intricate cloth configurations and deformation patterns from data. Therefore, we propose a diffusion-based generative approach for both perception and dynamics modeling. Specifically, we formulate state estimation as reconstructing full cloth states from partial observations and dynamics modeling as predicting future states given the current state and robot actions. Leveraging a transformer-based diffusion model, our method achieves accurate state reconstruction and reduces long-horizon dynamics prediction errors by an order of magnitude compared to prior approaches. We integrate our dynamics models with model predictive control and show that our framework enables effective cloth folding on real robotic systems, demonstrating the potential of generative models for deformable object manipulation under partial observability and complex dynamics.
\vspace{-5pt}
\end{abstract}
\keywords{Deformable Object Manipulation, Dynamics Model Learning, State Estimation, Generative Models, Cross-Embodiment Generalization
} 


%% file: text/010_introduction.tex
\section{Introduction}
Textile deformable objects, such as clothing, are ubiquitous in daily life. Yet, manipulating these objects is a long-standing challenge in robotics~\citep{Longhini2024unfolding, yin2021modeling}, due to their complex geometric structures and dynamics. Effective cloth manipulation requires accurately estimating the state of cloth despite severe self-occlusions, as well as reasoning over its complex, continuous dynamics to optimize actions. These difficulties highlight the need for advancements in both (i) state estimation and (ii) dynamics modeling to enable robust robotic cloth manipulation.

State estimation for cloth is particularly challenging due to frequent self-occlusions arising from its highly deformable structure. While humans intuitively infer full object shapes from partial observations using prior experience, most existing methods are unable to fully capture the complex mapping between highly partial observations and high-dimensional object states \citep{chi2021garmentnets, Huang2022MeshbasedDW, wang2023trtm}. A promising direction is to develop perception models that can \textit{``imagine''} full states from partial observations by leveraging extensive prior experience, akin to human reasoning.

Modeling cloth dynamics poses another significant challenge due to its highly nonlinear nature. Current approaches typically represent cloth using particle- or mesh-based structures and model their interactions with graph neural networks (GNNs) \citep{zhang2024adaptigraph, Huang2022MeshbasedDW, li2018learning, he2025learning}. GNNs offer advantages in data-scarce domains through spatial equivariance and locality, but they scale inefficiently with the number of graph nodes \citep{rong2020self}. Moreover, the locality inherent to graph structures often limits their ability to capture long-range dependencies, which is crucial for accurate dynamics modeling. 

In this work, we formulate state estimation and dynamics prediction as conditional generation processes. State estimation reconstructs full states from partial observations, while dynamics prediction generates future states conditioned on the current state and robot actions. To model these complex high-dimensional mappings, we employ diffusion-based models, inspired by their recent successes in capturing complex data distributions in computer vision~\citep{ma2024latte, liu2023zero}, science~\citep{ruhling2023dyffusion}, and robotics~\citep{chi2023diffusionpolicy}. We hypothesize that diffusion models with scalable architecture (\eg, Transformer~\citep{Vaswani2017attention}) can enable accurate state reconstruction and dynamics modeling. 


Building on these insights, we introduce \ours{}, a unified framework that integrates a Diffusion Perception Model (DPM), a Diffusion Dynamics Model (DDM), and model predictive control for cloth manipulation. Conceptually, DPM leverages diffusion models and Transformers to reconstruct full cloth states from sparse and occluded RGB-D observations, while DDM predicts long-horizon dynamics conditioned on current states and actions. Trained on a large-scale cloth interaction dataset with 500K transitions in simulation and evaluated in both simulation and real-world, our models achieve substantial performance gains: DPM achieves superior performance compared to prior approaches in cloth state estimation, and DDM reduces long-horizon prediction error by an order of magnitude compared to GNN-based baselines. With an embodiment-agnostic action representation, our framework can be deployed on both parallel grippers and dexterous hands. Real-world experiments demonstrate superior manipulation performance over previous approaches, highlighting the potential of generative modeling in deformable object manipulation.

%% file: text/020_related_work.tex
\vspace{-0.05cm}
\section{Related Work}
\vspace{-0.05cm}

\textbf{Deformable Object Manipulation.}
Manipulating deformable objects such as garments remains a long-standing challenge in robotics, due to their high-dimensional state space and complex, nonlinear dynamics. Model-free approaches, including reinforcement learning (RL)~\citep{pmlr-v87-matas18a,jan2020rl} and imitation learning (IL)~\citep{9981402,fu2024mobile, chi2023diffusionpolicy, Ze2024DP3, pertsch2025fast}, learn direct observation-to-action mappings through end-to-end training. However, these methods struggle with precise shape control due to the lack of explicit dynamics reasoning. Model-based approaches require accurate state estimation~\citep{Chen2021AbInitio, shi2022robocraft, shi2023robocook, ai2024robopack,huang2024rekep, longhini2024adafold}, which is highly challenging with partial observations.
Further, learning dynamics models demands extensive training data to cover large state and action spaces. Thus, we propose to learn expressive generative models for state estimation and dynamics modeling using large-scale simulation data.


\textbf{Learning-Based Dynamics Models.}
Learning-based dynamics models \cite{ai2025review} predict state transitions from interaction data, where the choice of state representation is crucial. Pixel-based models view the problem as action-conditioned video prediction~\citep{10.1007/s10514-021-10001-0, pmlr-v155-yan21a}, but they are often sample-inefficient, vulnerable to occlusions, and lack physical realism for contact-rich scenarios~\citep{yang2023learning}. Structured representations, such as particles or meshes, provide stronger physical priors and are typically coupled with graph neural networks (GNNs) that perform inference via message passing~\citep{ai2024robopack, zhang2024adaptigraph, Huang2022MeshbasedDW, shi2022robocraft, shi2023robocook, he2025learning}. While sample-efficient, GNNs often struggle with scalability and long-range interactions. In contrast, we find diffusion models offer greater expressiveness and scalability, enabling accurate dynamics prediction from large-scale data and improving modeling of deformable object behavior.


\textbf{Diffusion Models.}
Diffusion models~\citep{ho2020denoising}, a class of generative models with expressive capability of capturing complex, high-dimensional data distributions precisely, have emerged as a powerful paradigm and been applied across diverse domains, including generation of images~\citep{Rombach_2022_CVPR, Peebles2022DiT}, videos~\citep{wang2023lavie, ma2024latte}, 3D shapes~\citep{poole2022dreamfusion}, as well as robot policy learning~\citep{chi2023diffusionpolicy, Ze2024DP3} and world modeling~\citep{alonso2024diffusionworldmodelingvisual, ding2024diffusion}. In this work, we adapt diffusion models for deformable objects manipulation, leveraging their superior data distribution modeling capability for (i) estimating full cloth configurations from partial point cloud observations, and (ii) modeling state transitions to enable accurate future prediction and model-based planning for cloth manipulation.

%% file: text/030_method.tex
\section{Method}
\label{sec:method}

\subsection{Overview}

\begin{figure*}[!t]
    \centering
    \begin{tabular}{cc}
        \includegraphics[width=0.95\textwidth]{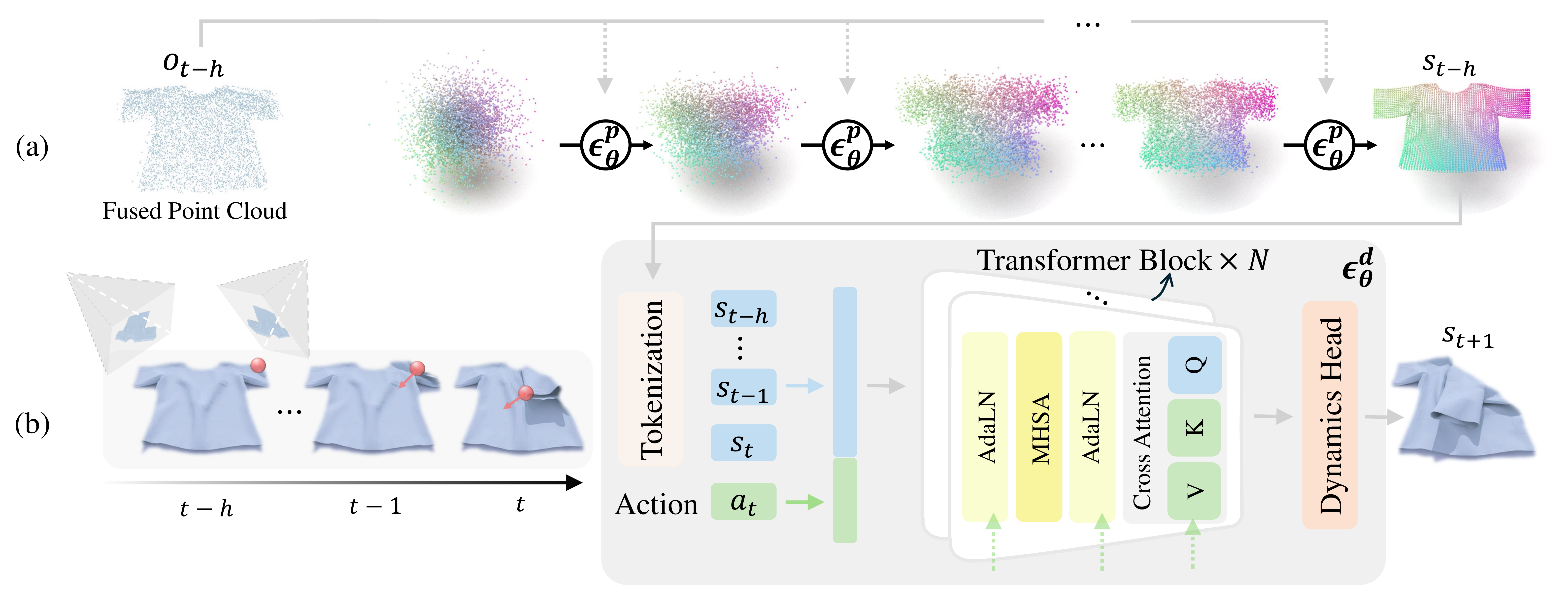} \vspace{-5pt}
    \end{tabular}
    \caption{\newtext{\textbf{Overview.} \textbf{(a) Perception:} Our Diffusion Perception Model (\oursperception) reconstructs the full cloth state from a partial point cloud. Using a denoising process parameterized by $\boldsymbol{\epsilon_{\theta}^p}$, \oursperception refines the cloth state over $K$ denoising steps, starting from random noise. \textbf{(b) Dynamics Prediction:} Our Diffusion Dynamics Model (\oursdynamics) generates future cloth states based on the current estimated state and robot actions, using a transformer-based architecture. }
    }
    \vspace{-0.5cm}
    \label{fig:method_pipeline}
\end{figure*}

We address the challenge of manipulating cloth with significant self-occlusions into target configurations. Our problem formulation comprises three key spaces: observation space $\O$, state space $\S$, and action space $\A$. The objective is to learn two essential components: a state estimator $g: \O \rightarrow \S$ and a transition function $\T: \S \times \A \rightarrow \S$ for model-based control.

At each timestep, the system processes multiview RGB-D observations $o_t \in \O$, represented as $o_t = \{I_t^0, I_t^1, \ldots, I_t^{l-1}\}$ with $l$ camera views, to estimate the cloth's 3D state $s_t \in \S$ given canonical state of the template mesh $s_c$. The state of the cloth is defined by a mesh $s_t = \{V_t, E_t\}$, where $E_t$ represents the invariant edge connectivity and $V_t \in \mathbb{R}^{N_v \times 3}$ denotes the positions of vertices in 3D space where $N_v$ denotes the number of vertices. We propose that generative models can effectively infer unobserved patterns in partial RGB-D observations, enabling robust state estimation.

Given the estimated state, a learned dynamics model $f$ predicts the future state $s_{t+1} \in \S$ based on state history $s_{t-i:t} \in \S$ and planned action $a_t \in \A$. This dynamics model is integrated with model-predictive control to optimize action sequences for achieving the target state $s_g$:
\setlength{\abovedisplayskip}{2pt}
\setlength{\belowdisplayskip}{2pt}
\begin{align*}
(a_0, ..., a_{H-1}) = \argmin_{a_0, ..., a_{H-1} \in \A{}} \J{}(\T(s_0, (a_0, .., a_{H-1})), s_g)
\end{align*}

\subsection{State Estimation}
We first address the problem of inferring cloth configurations from partial observations. 
Despite using four multi-view RGB-D cameras, severe self-occlusions make accurate state estimation infeasible. Inspired by the human ability to infer hidden object states from partial views, we propose using diffusion models to generate full cloth configurations from limited observations.

\textbf{Conditional Diffusion Process.} 
We formulate cloth state estimation as a conditional denoising diffusion process, using the object point cloud as the conditioning input. Conditioning on point clouds helps minimize the sim-to-real gap due to their nature as a mid-level visual representation and maintains geometric invariance~\citep{pmlr-v155-chen21f, ai2023invariance}.

Specifically, we model the conditional distribution $p(s|s_c, \mathbf{e}_\text{pc})$ using standard denoising diffusion probabilistic model (DDPM)~\citep{ho2020denoising}, where $s_c$ represents the state of the canonical cloth mesh and $\mathbf{e}_\text{pc}$ denotes the embedding of the conditional point cloud. To get point cloud embedding, we partition the point cloud into patches by first sampling M center points using farthest point sampling (FPS) and performing K-Nearest Neighbors (KNN) clustering. Then each resulting patch is processed through a PointNet~\citep{qi2016pointnet} to obtain its embedding representation $\mathbf{e}_\text{pc} \in R^{B \times M \times D_1}$. where $B$ is the batch size and $D_1$ is the dimension of the point cloud embedding.

In the forward process, starting from the initial state $s_0$, gaussian noise is gradually added at levels $t \in \{1, ..., T\}$ to get noisy state as:$s_t = \sqrt{\bar{\alpha}_t}s_0 + \sqrt{1-\bar{\alpha}_t}\boldsymbol{\epsilon}$
\label{eqn:diffusion_forward}
, where $\boldsymbol{\epsilon} \sim \mathcal{N}(0,I)$, $\bar{\alpha}_t := \prod_{s=1}^t 1-\beta_s$, and $\{\beta_1,\ldots,\beta_T\}$ is the variance schedule of a process with $T$ steps. In the reverse process, starting from a noisy state $s_t$ sampled from the normal distribution, the conditional denoising network $\epsilon^{p}_{\theta}$ gradually denoises from $s_t$ to $s_{t-1}$ and finally constructs $s_0$.
 
\textbf{Model Architecture.}
We adopt vanilla Vision Transformer (ViT) architecture~\citep{dosovitskiy2020vit} as our backbone, which has been shown to be highly scalable in image and video generation~\citep{Peebles2022DiT, ma2024latte}.
The model takes a point cloud and a canonical template mesh as input, in addition to the noisy mesh state that requires denoising. We detail our network architecture and training objective below.

\textbf{Tokenization.} We tokenize the input mesh as non-overlapping vertex patches in canonical space. We first use farthest point sampling (FPS) to sample a fixed number of points as patch centers $C \in \mathbb{R}^{N \times 3}$. To patchify the mesh vertices, we use the $N$ centers obtained from FPS to construct a Voronoi diagram in the 3D points space. This tessellation divides the point cloud into N distinct regions, where each region contains all points closer to its associated center than to any other center. Each Voronoi cell is treated as a distinct patch, encompassing a local neighborhood of points which will then go through a PointNet~\citep{qi2016pointnet} layer for feature extraction. 

\textbf{Conditioning.} Following the tokenization process, the input token is directly subjected to a sequence of transformer blocks for processing. To effectively condition the point cloud embedding, we adopt two approaches. First, the conventional layer normalization is replaced with an adaptive layer normalization (AdaLN)~\citep{xu2019understanding} to better incorporate conditional information, which modulates the normalization parameters based on the point cloud condition embedding for effective feature modulation.
Then, we incorporate conditional information through a cross-attention layer positioned after the multi-head self-attention (MHSA). In this cross-attention operation, the hidden states $x$ serve as the query vector, while the conditional information acts as both the key and value vectors. The computation proceeds as $x = \text{CrossAttention}(W_Q^{(c)}x, W_K^{(c)}\textbf{e}_{pc}, W_V^{(c)}\textbf{e}_{pc}) $\label{eq:h4}
where $W^{(c)}$ are learnable parameters, enabling effective conditioning during the learning process.

\textbf{Decoding.} Finally, the decoding process transforms the hidden states $x$ into 3D vertex coordinates through a two-stage process. First, we employ distance-weighted interpolation to upsample the hidden states, where interpolation weights are computed from canonical-space distances between vertices and their corresponding patch centers. This operation produces an intermediate representation $x \in \mathbb{R}^{B \times N_v \times D_2}$. A Multi-Layer Perceptron (MLP) then maps this representation to the final output $x_{out} \in \mathbb{R}^{B \times N_v \times 3}$, yielding the predicted noise added onto the 3D coordinates for each vertex during the diffusion forward process. Details of our model are presented in \appendref{appendix:model_details}.

\textbf{Training.}
We train the denoising model $\epsilon^{p}_{\theta}(s^{(k)} | s_c, \textbf{e}_{\text{pc}})$ to minimize the loss:
$$\mathcal{L}(\theta) = \mathbb{E}_{s, s_c, \mathbf{e}_{\text{pc}} \sim p_{\text{data}}} \left[ \left\| \epsilon - \epsilon^{p}_\theta\left(\sqrt{1-\beta^{(k)}}s + \sqrt{\beta^{(k)}}\epsilon \middle| s_c, \mathbf{e}_{\text{pc}}\right) \right\|^2 \right]$$
where $\epsilon \sim \mathcal{N}(0, I)$ and $\beta^{(k)} \in \mathbb{R}$ are $K$ different noise levels for $k \in [1, K]$. Details of the training process are available in \appendref{appendix:training_details}.

\subsection{Dynamics Prediction}
Given the estimated state, the goal of dynamics prediction is to reason about future states of the cloth given robot actions. \newtext{We extend our state estimation architecture to model dynamics by modifying the condition input to incorporate robot actions and enhancing the temporal modeling capability with additional temporal attention layers. The remaining components, including tokenization, training objective, and decoding of the model, are identical to those in the state estimation framework.}

\textbf{Conditional Diffusion Process.} 
To learn the conditional posterior distribution $p(s_{t+1:t+j+1}|a_t,s_{t-i:t})$, we parameterize it using diffusion models. Here, $a_t$ represents the robot action, $s_{t-i:t}$ denotes the historical states, and $s_{t+1:t+j+1}$ is the $j$ frame future states to be predicted at timestep $t$. Following prior work ~\citep{zhang2024adaptigraph, yang2023learning}, we heuristically set $i=3$ and $j=5$. The diffusion reverse process construct $s_{t}$ conditioned on history frames and action by gradually denoising from a normal distribution with the denoising network $\epsilon_{\theta}^d$. \newtext{Since we use delta end-effector position as action representation, to effectively encode the action space, we employ a Fourier feature-based embedding following NeRF~\citep{mildenhall2021nerf} to represent continuous spatial information, with detailed formulation in~\appendref{appendix:model_details}.} 

\subsection{Model-Based Planning}
We integrate our diffusion dynamics model with Model Predictive Control (MPC) for robotic cloth manipulation. Given a current cloth state sequence $s_{t-i:t} \in \S$ and target state $s_g$, we optimize an action sequence $\{a_t\}_{t=0}^{T-1}$ over horizon $T$ by minimizing:
\begin{equation}
\min_{\{a_t\}_{t=0}^{T-1}} \phi\left(s_T, s_g\right) + \sum_{t=0}^{T-1} \ell\left(s_t, a_t\right),
\end{equation}
where $\phi$ combines weighted mean squared error (MSE) and chamfer distance (CD), and $\ell$ enforces action smoothness. We use Model Predictive Path Integral (MPPI)~\citep{7487277} for sampling-based optimization. Actions are defined as relative end-effector displacements applied to a selected cloth grasp point. To improve planning efficiency, we introduce an informed action sampling strategy and a probabilistic grasp point selection mechanism. Specifically, the grasp point is selected using a temperature-controlled softmax distribution based on vertex displacements between the current and target states, while action sampling is guided by a weighted direction computed from high-displacement vertices. After each action, the robot updates its state estimate using DPM before replanning. Refer to \appendref{appendix:planning_details} for details on the planning algorithm and hyperparameters.

%% file: text/040_experiments.tex
\section{Experiments}
\label{others}


We investigate three key research questions: \textbf{(1)} How effectively does the Diffusion Perception Model handle self-occlusions inherent in cloth manipulation? \textbf{(2)} How does the Diffusion Dynamics Model improve dynamics prediction compared to prior approaches? \textbf{(3)} How do these enhanced perception and dynamics models translate to overall system performance?
We study these questions by evaluating state estimation accuracy (Section~\ref{sec::state_est}), assessing dynamics modeling performance (Section~\ref{sec::dynamics}), and real-world deployment across two system setups(Section~\ref{sec::planning}).

\subsection{Experiments Setup}
We evaluate our method in both simulation and real-world environments. Specifically, we use SAPIEN~\citep{SAPIEN} as the simulation platform for data collection and training, and demonstrate effective sim-to-real transfer in the real-world setting. Additional details of the pipeline implementation and experimental setup are provided in \appendref{sec:experiment_setup}.

\subsection{State Estimation Results}\label{sec::state_est}

\begin{figure}[t]
    \centering
    \begin{tabular}{cc}
    \includegraphics[width=1.0\textwidth]{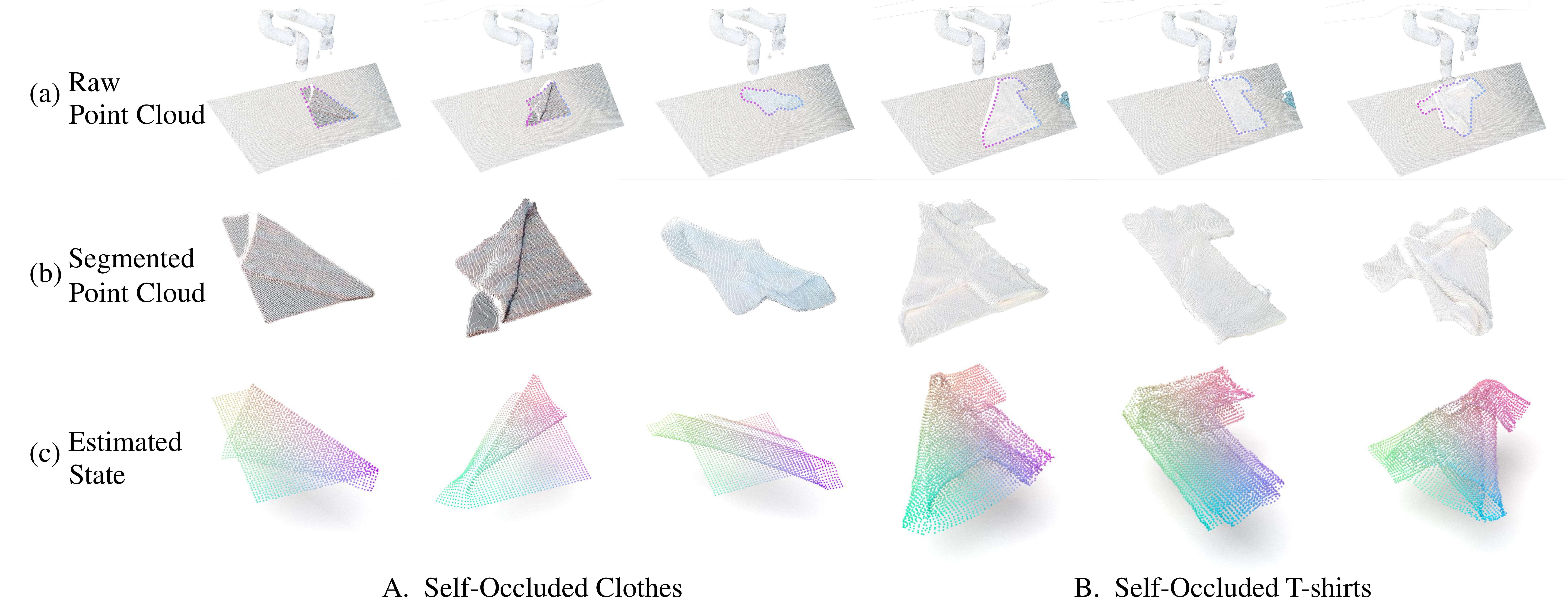}
    \end{tabular}
    \caption{\textbf{Qualitative results on state estimation.}  Row (a) shows raw point cloud of the work space. Row (b) shows the segmented point clouds of \textbf{real-world} clothes, all of which are highly crumpled. Row (c) shows the predicted cloth states.
    }
    \vspace{-0.5cm}
    \label{fig:state_est_qualitative}
\end{figure}

\textbf{Baselines.}
We compare our perception module against four baselines: \textbf{\garmentnets}~\citep{chi2021garmentnets} which formulates cloth pose estimation problem as a shape completion task in the canonical space; \textbf{\medor}~\citep{Huang2022MeshbasedDW} which improves \garmentnets by introducing test-time fine-tuning for mesh refinement; \textbf{\trtm}~\citep{wang2023trtm} which employs a template-based approach for explicit mesh reconstruction; and \textbf{\dpmwoddpm}, an ablated version of our model that retains the original architecture but without diffusion training. 
These baselines cover both optimization-based and non-optimization-based prior works on cloth pose estimation, along with ablation studies for our model.

\textbf{Results.}
We evaluate our method and baselines in both simulation and the real world using MSE, CD, and Earth Mover’s Distance (EMD). The results are presented in~\tabref{tab:state_estimation}.
In the T-shirt setting, \trtm and \dpmwoddpm greatly outperform \garmentnets and \medor, demonstrating that the topological information provided by the template cloth mesh significantly enhances the perception capabilities. Leveraging the cloth modeling prior during the learning process, \trtm demonstrates better performance compared to \dpmwoddpm. Our approach achieves further performance gains over both \trtm and \dpmwoddpm, highlighting the significant contributions of diffusion models to the task. We provide qualitative results in \figref{fig:state_est_qualitative}.

\begin{table*}[h]
\centering
\begin{threeparttable}
   \setlength{\tabcolsep}{5pt}
   \resizebox{0.95\textwidth}{!}{
   \begin{tabular}{
   >{\centering\arraybackslash}m{0.8cm} 
   >{\centering\arraybackslash}m{2.4cm} 
   c@{\hspace{3pt}}c@{\hspace{3pt}}c@{\hspace{8pt}}c@{\hspace{3pt}}c}
   \toprule
   \multirow{2}{*}{\textbf{Category}} & \multirow{2}{*}{\textbf{Method}} & \multicolumn{3}{c}{\textbf{Simulation}} & \multicolumn{2}{c}{\textbf{Real World}} \\
   \cmidrule(lr){3-5} \cmidrule(lr){6-7}
   & & MSE $\downarrow$  ($10^{-1}$) &  CD $\downarrow$ ($10^{-1}$) &  EMD $\downarrow$ ($10^{-1}$) & CD $\downarrow$ ($10^{-1}$) &  EMD $\downarrow$ ($10^{-1}$) \\
   \midrule
   \multirow{3}{*}{Cloth} 
   & \trtm\citep{wang2023trtm} & 5.07 ± 0.22 & 2.67 ± 0.61 & 1.65 ± 0.71 & 1.85 ± 0.15 & 0.86 ± 0.23\\
   & \dpmwoddpm & 5.44 ± 0.41 & 2.17 ± 0.19 & 1.61 ± 0.45 & 1.72 ± 0.22 & 0.78 ± 0.33 \\
   & \textbf{\oursperception} & \cellcolor{orange!20}\textbf{2.32} ± 0.21 & \cellcolor{orange!20}\textbf{1.95} ± 0.25 & \cellcolor{orange!20}\textbf{1.48} ± 0.47 & \cellcolor{orange!20}\textbf{1.13} ± 0.25 & \cellcolor{orange!20}\textbf{0.54} ± 0.49 \\
   \midrule
   \multirow{5}{*}{T-shirt}
   & \garmentnets\citep{chi2021garmentnets} & 18.6 ± 1.35 & 6.23 ± 0.79 & 2.79 ± 0.64 & 7.18 ± 0.51 & 2.86 ± 0.46\\
   & \medor\citep{Huang2022MeshbasedDW} & 21.0 ± 1.54 & 6.87 ± 0.95 & 2.24 ± 0.29 & 5.01 ± 0.48 & 2.49 ± 0.32 \\
   & \trtm\citep{wang2023trtm} & 6.30 ± 0.45 & 5.15 ± 0.96 & 2.15 ± 0.29 & 3.18 ± 0.44 & 1.99 ± 0.29 \\
   & \dpmwoddpm & 9.12 ± 0.57 & 5.56 ± 0.63 & 1.99 ± 0.62 & 2.34 ± 0.37 & 1.91 ± 0.33  \\
   & \textbf{\oursperception} & \cellcolor{orange!20}\textbf{2.76} ± 0.19 & \cellcolor{orange!20}\textbf{3.22} ± 0.41 & \cellcolor{orange!20}\textbf{1.95} ± 0.56 & \cellcolor{orange!20}\textbf{2.17} ± 0.28 & \cellcolor{orange!20}\textbf{1.88} ± 0.61 \\
   \bottomrule
   \end{tabular}
   }
\end{threeparttable}
\caption{\textbf{Quantitative results on state estimation.} We report estimation errors in both simulated and real-world scenarios, with 95\% confidence intervals. Lower values indicate better performance.}
\label{tab:state_estimation}
\vspace{-0.5cm}
\end{table*}


\subsection{Dynamics Prediction Results} 
\label{sec::dynamics}

\textbf{Baselines.}
\newtext{We evaluated our diffusion dynamics models against three baseline approaches: a \textbf{\gnn}-based method~\citep{zhang2024adaptigraph} which is the most widely adopted approach for modeling dynamics; an \textbf{Analytical Simulator} specifically for configurations using the \oursperception's output; and an ablated version of our model with dynamics module trained directly with MSE loss supervision termed \textbf{\dpmwoddpm}. For each baseline model, we analyze the MSE across different timesteps on clothes and T-shirts.}

\textbf{Results.}
Our evaluation compares the proposed approach against three baselines using MSE across two experimental scenarios: (1) using ground truth states from the simulator and (2) using perception states estimated by \oursperception. The second scenario, which includes a direct comparison with \simulator, demonstrates the robustness of our method to noisy states that typically degrade the performance of the analytical simulator.

\begin{figure}[t!]
    \centering
    \includegraphics[width=1\textwidth]{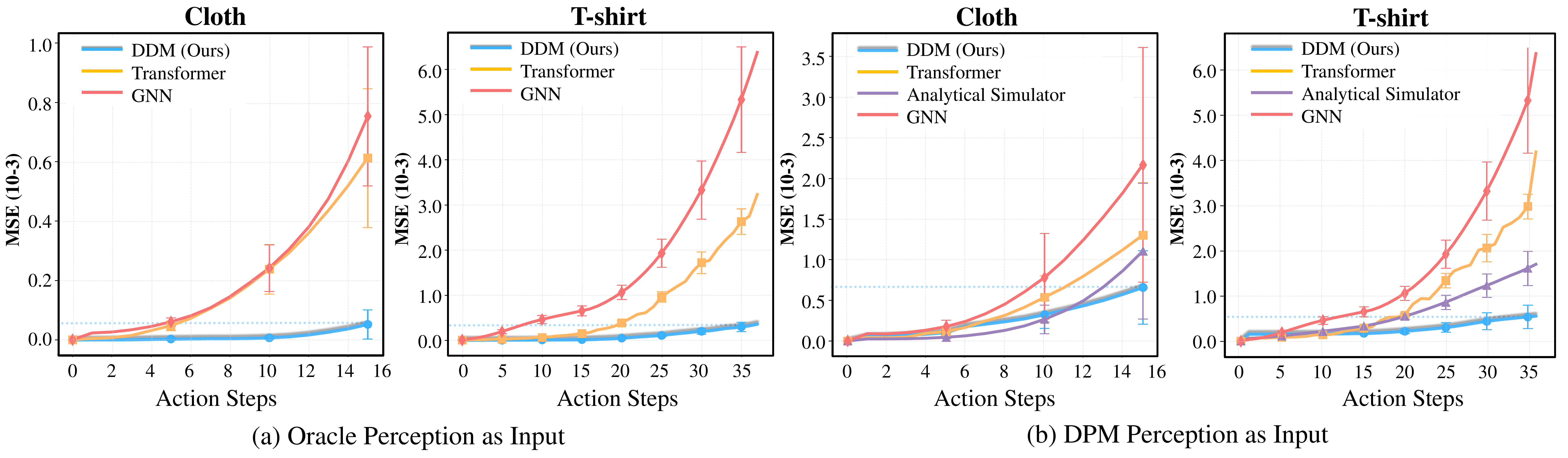}
    \caption{\textbf{Long-horizon dynamics prediction error over time.} MSE in dynamics prediction over time under two scenarios: (a) using oracle simulation states, and (b) using DPM perception estimates, evaluated on clothes and T-shirts. Error bars represent 95\% confidence intervals.}
    \vspace{-0.5cm}
    \label{fig:dynamics_mse}
\end{figure}
\begin{figure}[t]
    \centering
    \includegraphics[width=1\linewidth]{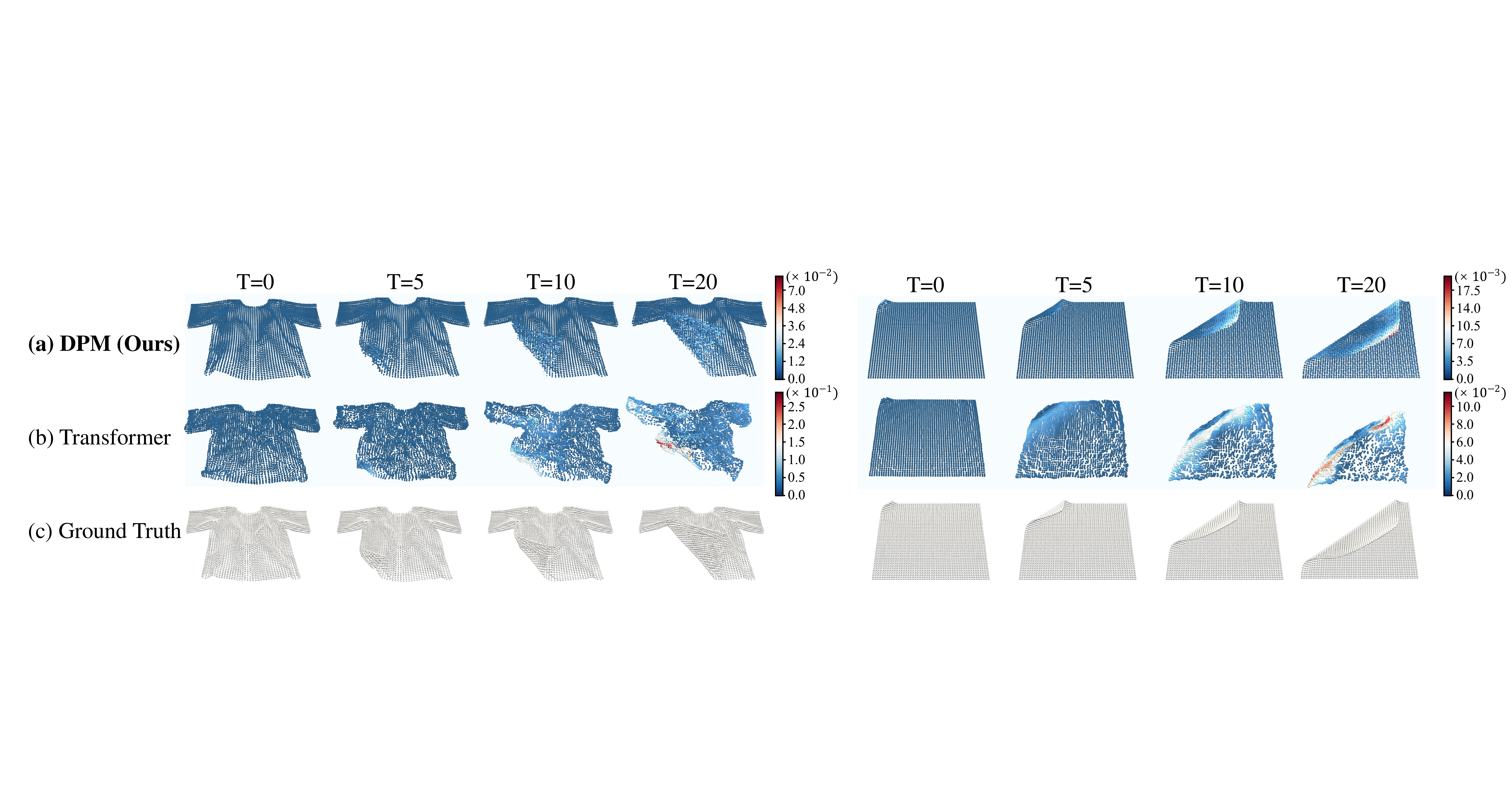}
    \caption{\textbf{Qualitative results on dynamics prediction.} Visualization of predicted clothes configurations with each vertex color-coded by its point-wise L2 error from ground truth.}
    \label{fig:dynamics_qualitative_results}
    \vspace{-0.5cm}
\end{figure}

Error analysis over time in \figref{fig:dynamics_mse} shows that DDM consistently outperforms all baselines. \gnn exhibits the weakest performance, particularly for complex objects like T-shirts. \dpmwoddpm improves over \gnn by leveraging transformer architectures, but still suffers from error accumulation. In contrast, \oursdynamics achieves the lowest MSE across all timesteps with minimal temporal error accumulation, benefiting from the diffusion model’s expressive distribution modeling. Qualitative results in \figref{fig:dynamics_qualitative_results} further highlight the physical plausibility of \oursdynamics's predictions.

When using estimated states with perception noise, we introduce \simulator as an additional baseline. Although \simulator initially achieves low error on cloth objects, it is highly sensitive to inconsistent inputs, leading to rapid error accumulation and worse long-horizon performance than \oursdynamics. This degradation is even more pronounced for T-shirt objects due to their complex topology. Under these realistic conditions, both \gnn and \ddmwoddpm exhibit even larger gaps, demonstrating that \oursdynamics provides the most robust and accurate dynamics predictions for planning with noisy observations. We provide additional quantitative results in \appendref{appendix:dynamics}.

\subsection{Real World Planning Results} 
\label{sec::planning}
\begin{wrapfigure}{r}{0.5\textwidth}
  \centering
  \footnotesize
  \setlength\tabcolsep{3pt}
  \renewcommand{\arraystretch}{1.05}
        
        \vspace{-0.4cm}
        \setlength\tabcolsep{2pt}
        \renewcommand{\arraystretch}{1.05}
        {\scriptsize
        \begin{tabular}{@{}c ccc ccc ccc@{}}
            \toprule
            \multirow{2}{*}{\textbf{Method}} & \multicolumn{3}{c}{\textbf{Cloth}} & \multicolumn{3}{c}{\textbf{T-shirt}} & \multicolumn{3}{c}{\textbf{Long-sleeve}} \\
            \cmidrule(lr){2-4} \cmidrule(lr){5-7} \cmidrule(lr){8-10}
             & \textbf{Self} & \textbf{Ext.} & \textbf{Comb.} & \textbf{Self} & \textbf{Ext.} & \textbf{Comb.} & \textbf{Self} & \textbf{Ext.} & \textbf{Comb.} \\
            \midrule
            GNN & 6/10 & 4/10 & 3/10 & 1/10 & 2/10 & 2/10 & 2/10 & 2/10 & 0/10 \\
            Ours & \textbf{9/10} & \textbf{8/10} & \textbf{6/10} & \textbf{9/10} & \textbf{7/10} & \textbf{6/10} & \textbf{7/10} & \textbf{6/10} & \textbf{4/10} \\
            \bottomrule
            \end{tabular}
        }
        \captionof{table}{\small \textbf{Quantitative results of real-world manipulation.} We repeat each scenario for 10 trials, with randomized initial and target states.}
        \vspace{1em}
        
        \label{tab:real_quantitative}
        
        {\scriptsize
        \begin{tabular*}{0.95\linewidth}{@{\extracolsep{\fill}}l@{\hspace{0.1em}}c@{\hspace{0.1em}}l|cc}
        \toprule
        \multicolumn{3}{l|}{\textbf{Method (Dynamics + Perception)}} & \textbf{Cloth SR}$\uparrow$ & \textbf{T-shirt SR}$\uparrow$ \\
        \midrule
        \rowcolor{orange!25} \textbf{DDM} &+& \textbf{DPM (Ours)} & \textbf{9/10} & \textbf{8/10} \\
        Transformer &+& DPM & 3/10 & 5/10 \\
        GNN &+& DPM & 6/10 & 1/10 \\
        DDM &+& Transformer & 5/10 & 3/10 \\
        Transformer &+& Transformer & 2/10 & 1/10 \\
        GNN &+& Transformer & 5/10 & 0/10 \\
        \bottomrule
        \end{tabular*}
        \captionof{table}{\small Success rates of system variants with different combinations of dynamics and perception modules.}
        \vspace{-0.3cm}
        \label{tab:success_rate}
        }
\end{wrapfigure}
\begin{figure}[t!]
    \vspace{-0.5cm}
    \centering
    \includegraphics[width=0.9\textwidth]{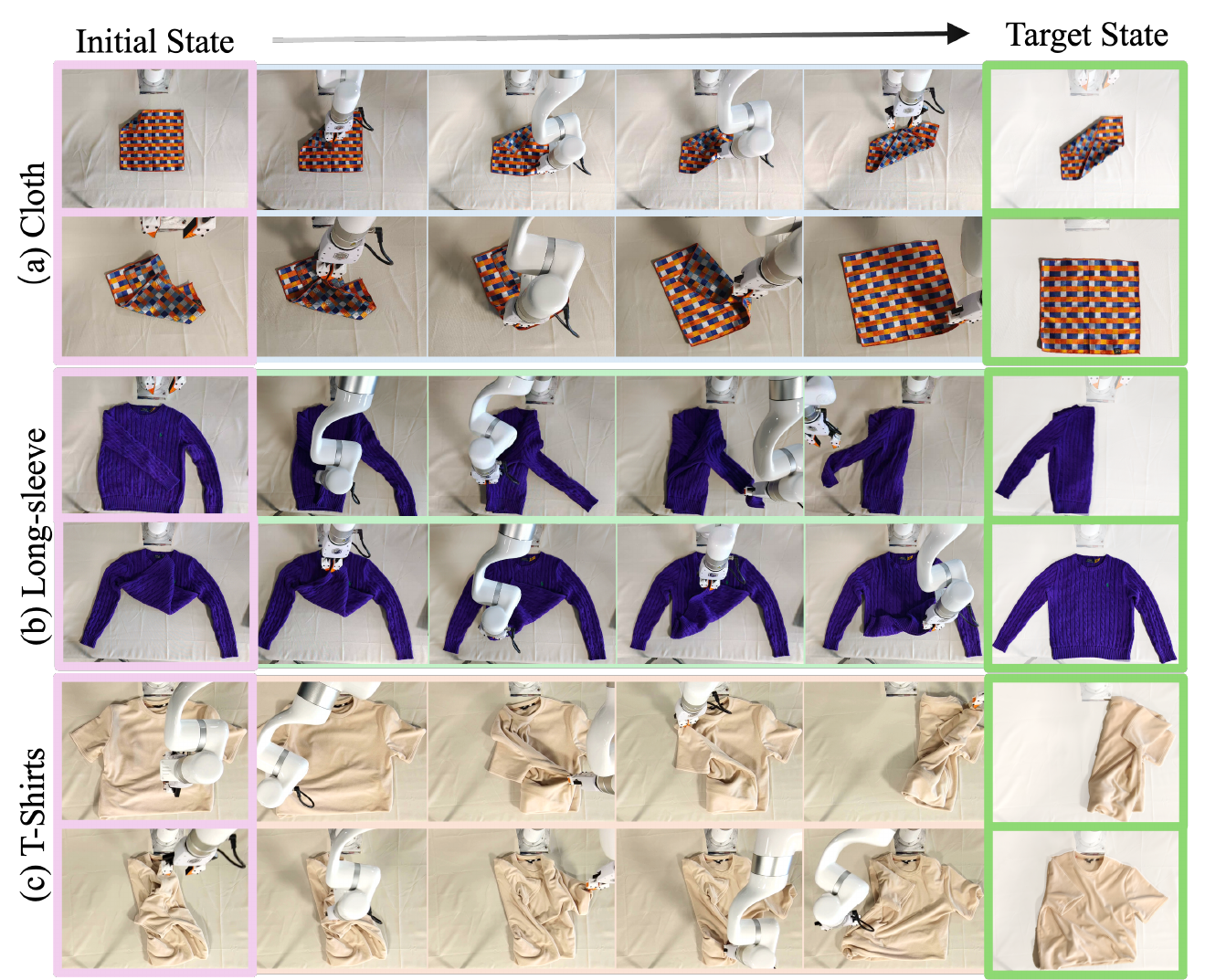}
    \caption{\textbf{Qualitative results on real-world cloth manipulation.} We evaluate our system on three types of garment: square clothes, long-sleeve shirts and T-shirts. For each garment, the first row corresponds to folding and the second row to unfolding.}
    \vspace{-0.5cm}
    \label{fig:real_qualitative}
\end{figure}
\begin{wrapfigure}{r}{0.55\textwidth}
\centering
\includegraphics[width=0.55\textwidth]{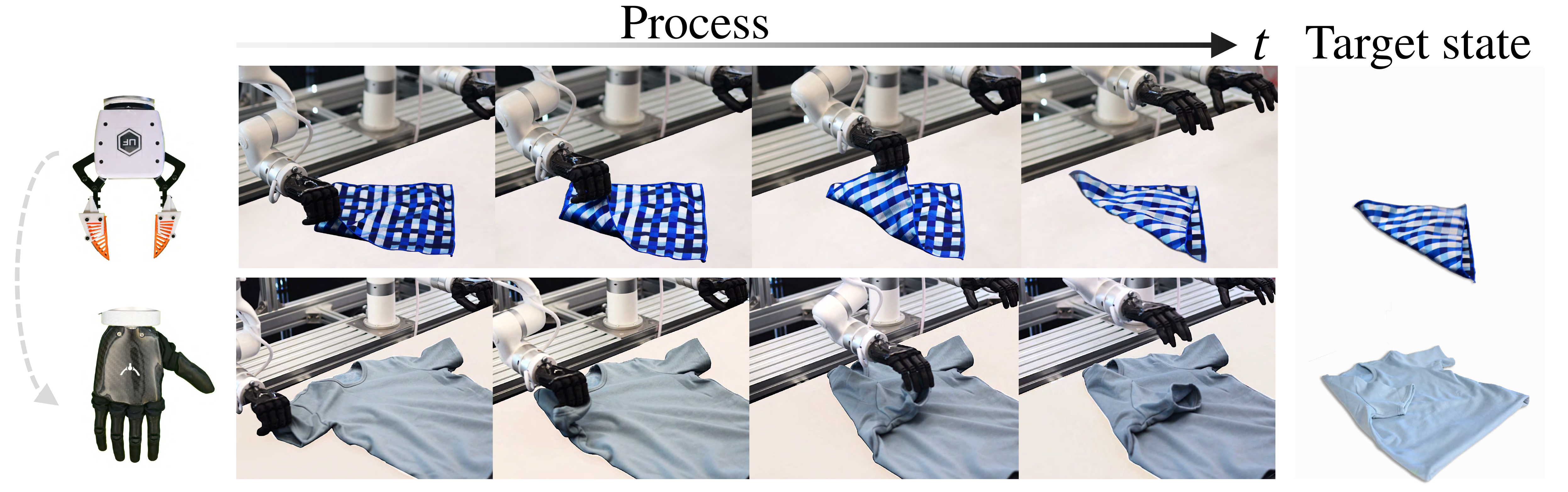}
\caption{\textbf{Cross-embodiment generalization results.}}
\label{fig:planning_qualitative_results_cross_embodiment}
\vspace{-0.2cm}
\end{wrapfigure}

\textbf{Comparative Analysis.} 
We demonstrate the seamless integration of \oursperception and \oursdynamics within an MPC framework for complex cloth folding tasks. Our approach is benchmarked against \gnn as a dynamics module. We categorize our tasks into folding and unfolding scenarios, all of which involve long-horizon challenges and require multi-step prediction, with three distinct occlusion types: self-occlusion, external occlusion by other objects (e.g., a robotic arm), and combined occlusion, which poses challenges for accurate perception in cloth manipulation. We report the success rates (SR) of the quantitative results in \tabref{tab:real_quantitative}. A trial is deemed successful if the geometric metric (EMD) is below a certain threshold (\appendref{appendix:planning}). Our method consistently outperforms \gnn across all occlusion scenarios. In simpler tasks, such as folding clothes, our model achieves an improvement of approximately 30\% points in SR. When manipulating more challenging objects, such as a dual-layer T-shirt where \gnn struggles to accurately model dynamics, our approach achieves up to a 50\% increase in SR.  Qualitative results are shown in \figref{fig:real_qualitative}, illustrating challenging initial and target configurations with severe self-occlusion, representing substantially more difficult setups than those considered in many prior works~\citep{Huang2022MeshbasedDW, zhang2024adaptigraph, 9981402, huang2024rekep}. 
By leveraging an embodiment-agnostic action space design, our approach enables effective cross-embodiment transfer. We demonstrate a successful transfer of the embodiment from a parallel gripper to a dexterous hand in \figref{fig:planning_qualitative_results_cross_embodiment}.

\textbf{Ablation.}
Ablation study results (\tabref{tab:success_rate}) demonstrate the critical contributions of both \oursperception and \oursdynamics to the system's overall performance, with their combination yielding significantly higher success rates than either component alone. For objects with simpler topology, such as cloth, accurate perception is most critical since the dynamics are relatively straightforward and easy to model. However, for objects with more complex topologies, having an accurate dynamics model becomes equally important for effective planning.


%% file: text/050_conclusion.tex
\vspace{-0.1cm}
\section{Conclusion} 
\label{sec:conclusion}
\vspace{-0.1cm}
We introduce \ours, a unified framework that tackles key challenges in state estimation and dynamics prediction in cloth manipulation with Transformer-based diffusion models. 
Our approach reconstructs full cloth configurations from partial RGB-D observations and predicts long-horizon dynamics with significantly lower error than prior GNN-based methods. 
Integrated with model-based control, it enables cloth manipulation in various scenarios, significantly outperforming existing approaches. Through extensive experiments, we demonstrate the potential of generative models for deformable object manipulation, paving the way for more robust and versatile robotic systems.

\section{Limitations}
One limitation of our method is the substantial computational cost associated with training large transformer-based diffusion models. 
Second, our experiments primarily focus on cloth manipulation; extending the framework to contact-rich rigid body tasks is a promising future direction. Given the generality of the model design, we expect feasible adaptation with suitable training data.
Additionally, our model currently lacks explicit uncertainty estimates. Incorporating uncertainty quantification into perception and dynamics models, and combining them with control methods with theoretical guarantees, could improve robustness in safety-critical scenarios. 

\section{Acknowledgments}
This research was funded by Hillbot Inc. Hao Su is the CTO for Hillbot and receives income. The terms of this arrangement have been reviewed and approved by the University of California, San Diego in accordance with its conflict of interest policies. We thank Xuhui Kang for his help with video production and editing and Zhan Ling for his helpful discussions.

%% file: text/060_appendices.tex
\appendix
\appendixpage

\section{Additional Results}



\subsection{Dynamics Prediction}
\label{appendix:dynamics}
\paragraph{Quantitative Results}
We provide additional quantitative results of forward dynamics prediction for both ground truth state and perception noisy input in~\tabref{tab:dynamics_quantitative_gt_input} and~\tabref{tab:dynamics_quantitative_perception_input}. Our dynamics model consistently outperforms all baselines across both evaluation scenarios. When using ground-truth states from the simulator as input, \oursdynamics{} achieves approximately 10× lower error than the best-performing baseline, highlighting the superiority of diffusion models in capturing long-horizon dynamics. When using estimated states from \oursperception{}, which introduces additional noise, \oursdynamics{} still achieves 2× lower error than all baselines, demonstrating that the diffusion-based training paradigm significantly enhances noise tolerance through its expressive data distribution modeling capacity.

\begin{figure}[h]
\centering
\begin{minipage}[t]{0.45\textwidth}
    \centering
    \scriptsize
    \begin{threeparttable}
    \begin{tabular}{p{0.5cm}|c|w{c}{1cm}w{c}{1cm}w{c}{1cm}}
    \toprule
    \textbf{Type} & \textbf{Method} & \textbf{MSE} $\downarrow$ & \textbf{CD} $\downarrow$ & \textbf{EMD} $\downarrow$ \\
    & & ($10^{-3}$) & ($10^{-2}$) & ($10^{-2}$) \\
    \midrule
     \multirow{3}{*}{Cloth}
    & \gnn & 0.75 $\pm$ 0.23  & 3.89 $\pm$ 0.80& 6.13 $\pm$ 2.96\\
    & \ddmwoddpm & 0.61 $\pm$ 0.23 & 1.85 $\pm$ 0.33& 5.47 $\pm$ 1.50\\
    & \textbf{\oursdynamics} & \cellcolor{orange!20} \textbf{0.05} $\pm$ 0.04 & \cellcolor{orange!20}\textbf{0.63} $\pm$ 0.28& \cellcolor{orange!20}\textbf{3.47} $\pm$ 0.60\\
    \midrule
    \multirow{3}{*}{T-shirt} 
    & \gnn & 6.36 $\pm$ 1.45& 8.57 $\pm$ 1.06& 7.89 $\pm$ 1.79\\
    & \ddmwoddpm & 3.22 $\pm$ 0.30& 2.80 $\pm$ 0.23& 7.57 $\pm$ 0.52\\
    & \textbf{\oursdynamics} & \cellcolor{orange!20} \textbf{0.35} $\pm$ 0.13 & \cellcolor{orange!20}\textbf{0.73} $\pm$ 0.07& \cellcolor{orange!20}\textbf{2.84} $\pm$ 0.47\\
    \bottomrule
    \end{tabular}
    \captionof{table}{\textbf{Quantitative results on dynamics prediction with ground truth input.}
    Errors represent a 95\% confidence interval.}
    \label{tab:dynamics_quantitative_gt_input}
\end{threeparttable}
\end{minipage}
\hfill
\begin{minipage}[t]{0.45\textwidth}
    \centering
    \scriptsize
    \begin{threeparttable}
    \begin{tabular}{p{0.5cm}|c|w{c}{1cm}w{c}{1cm}w{c}{1cm}}
    \toprule
    \textbf{Type} & \textbf{Method} & \textbf{MSE} $\downarrow$ & \textbf{CD} $\downarrow$ & \textbf{EMD} $\downarrow$ \\
    & & ($10^{-3}$) & ($10^{-2}$) & ($10^{-2}$) \\
    \midrule
    \multirow{3}{*}{T-shirt} 
    & \gnn & 6.36 $\pm$ 1.30& 8.88 $\pm$ 1.12& 8.29 $\pm$ 1.94\\
    & \ddmwoddpm & 4.18 $\pm$ 0.73& 4.26 $\pm$ 0.51& 7.93 $\pm$ 0.70\\
    & \textbf{\oursdynamics} & \cellcolor{orange!20} \textbf{0.55} $\pm$ 0.27 & \cellcolor{orange!20}\textbf{1.49} $\pm$ 0.13& \cellcolor{orange!20}\textbf{3.22} $\pm$ 0.47\\
    \midrule
    \multirow{3}{*}{Cloth}
    & \gnn & 2.17 $\pm$ 1.44 & 5.02 $\pm$ 0.90& 7.31 $\pm$ 4.65\\
    & \ddmwoddpm & 1.30 $\pm$ 0.65 & 2.27 $\pm$ 0.46& 7.06 $\pm$ 2.08\\
    & \textbf{\oursdynamics} & \cellcolor{orange!20} \textbf{0.66} $\pm$ 0.45 & \cellcolor{orange!20}\textbf{2.12} $\pm$ 0.54& \cellcolor{orange!20}\textbf{5.51} $\pm$ 1.03\\
    \bottomrule
    \end{tabular}
    \captionof{table}{\textbf{Quantitative results on dynamics prediction with perception input.} 
    Errors represent a 95\% confidence interval.}
    \label{tab:dynamics_quantitative_perception_input}
    \end{threeparttable}
\end{minipage}
\end{figure}

\begin{figure}[h]
\vspace{-0.7cm}
    \centering
    \includegraphics[width=0.9\linewidth]{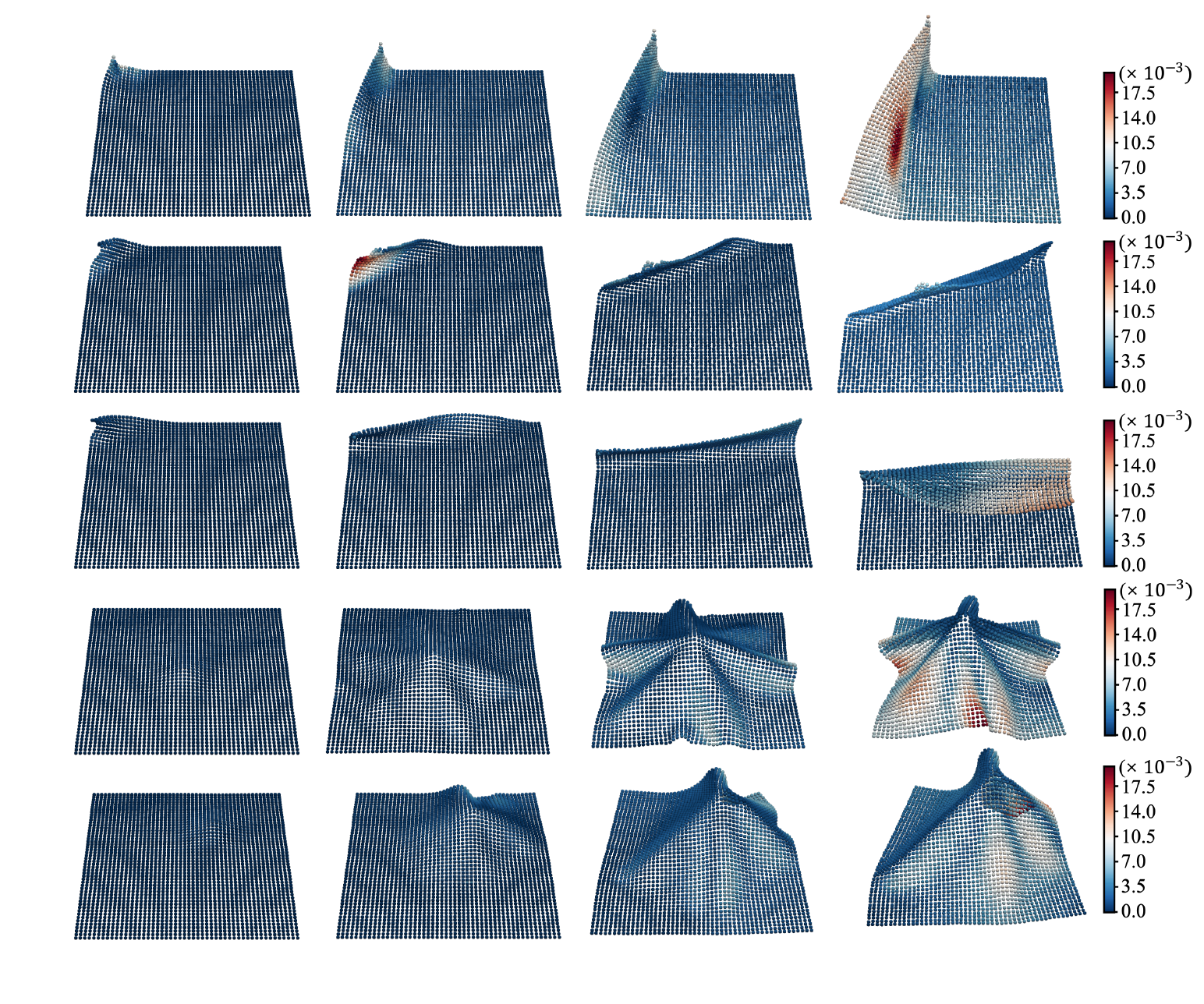}
    \vspace{-0.5cm}
    
    \caption{\textbf{Qualitative results on cloth dynamics prediction using DDM.}}
    \label{fig:cloth_qualitative_results_appendi}
    \vspace{-0.8cm}
\end{figure}

\begin{figure}[h]
    \centering
    \includegraphics[width=0.9\linewidth]{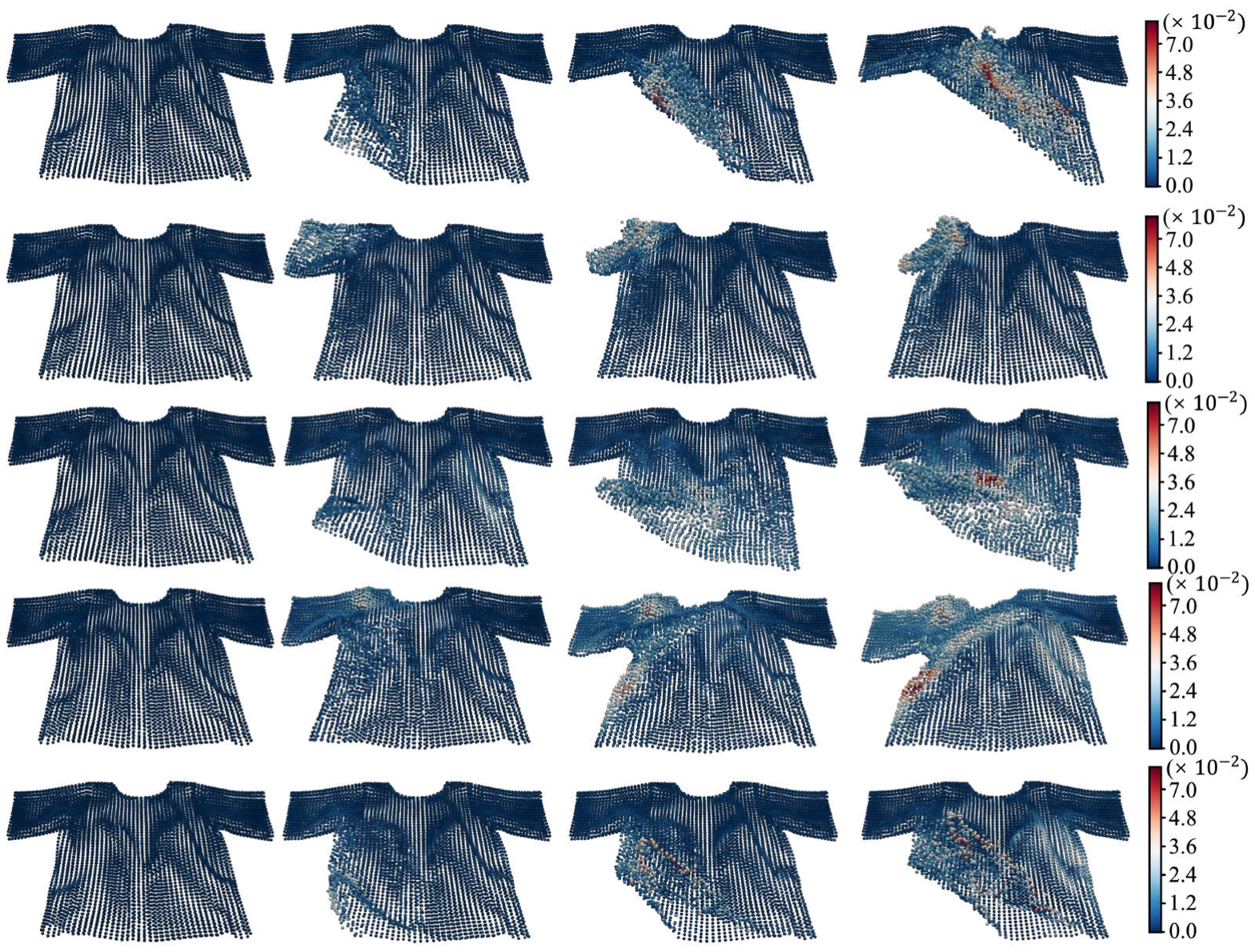}
    \caption{\textbf{Qualitative results on t-shirt dynamics prediction using DDM.}}
    \label{fig:tshirt_qualitative_results_appendix}
    \vspace{-0.4cm}
\end{figure}

\paragraph{Qualitative Results}
We present additional qualitative results for dynamics prediction in \figref{fig:cloth_qualitative_results_appendi} and \figref{fig:tshirt_qualitative_results_appendix}. Each row represents a predicted dynamics sequence. The results demonstrate the physical plausibility of the generated outputs.

\clearpage
\subsection{Real-world Planning}
\label{appendix:planning}
\begin{figure}[h]
     \centering
    \includegraphics[width=1\textwidth]{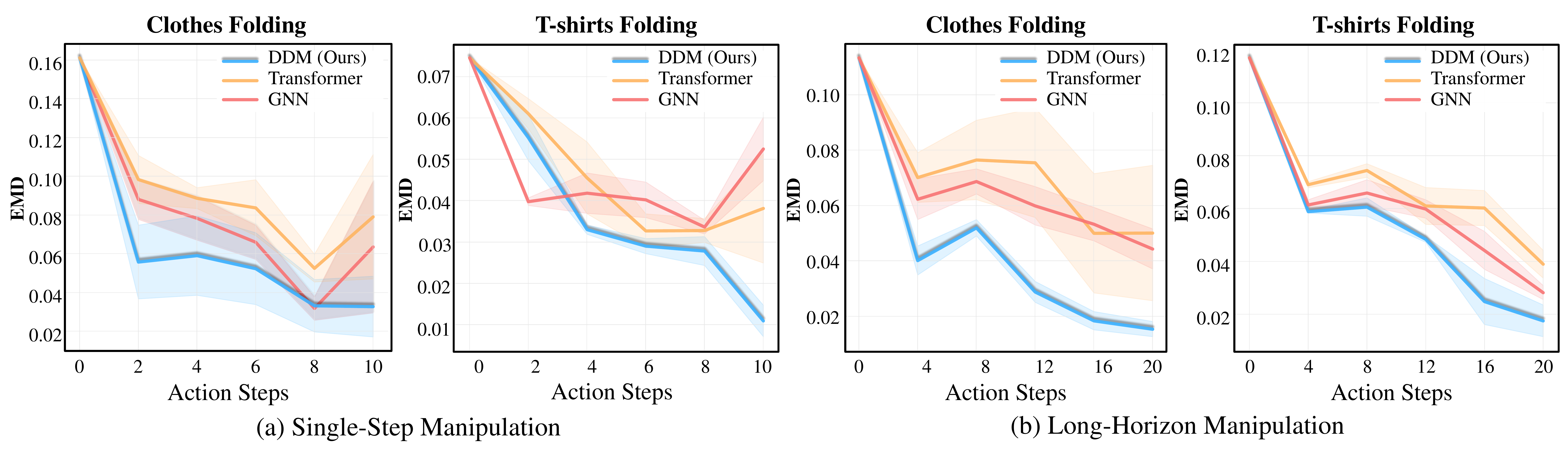}
    \caption{\textbf{Quantitative evaluation of planning performance.} Earth Mover's Distance (EMD) convergence during the planning stage, measured over 10 repeated trials with identical initial and target configurations. Errors represent 95\% confidence intervals. Our method outperforms baselines by achieving lower EMD values and faster convergence to goal states.}
    \vspace{-0.5cm}
    \label{fig:quant_planning_results}
\end{figure}

\paragraph{Additional Quantitative Results} We present quantitative results using the EMD metric, which measures the distance from the initial state to the target state in real-world planning scenarios in \figref{fig:quant_planning_results}.

 In single-step manipulation scenarios, our dynamics model exhibits superior performance across both object types. For cloth folding, our method consistently achieves lower EMD values with reduced confidence intervals, indicating enhanced prediction reliability compared to baseline approaches. This performance advantage is particularly evident in T-shirt folding, where topological complexity presents heightened challenges. While baseline methods, especially GNN, exhibit increased variance and elevated EMD values, our approach demonstrates consistent performance improvements throughout the planning horizon, suggesting enhanced handling of complex geometric relationships.

The multi-step scenarios, extending to 20 steps, further highlight our method's efficacy in long-horizon predictions. Our approach maintains significantly reduced EMD values with a consistent downward trajectory for both cloth and T-shirt manipulation tasks. The performance gap between our method and baselines becomes increasingly pronounced over extended horizons, particularly in T-shirt manipulation, where dual-layer structures introduce additional complexity. This sustained performance advantage in multi-step scenarios underscores our model's robust capability in mitigating error accumulation while maintaining prediction accuracy across extended planning sequences.


\begin{wrapfigure}{r}{0.55\textwidth}
\centering
    \includegraphics[width=0.5\textwidth]{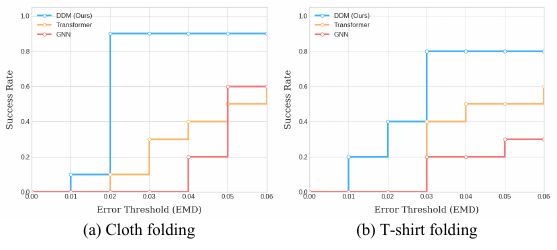}
    \captionsetup{aboveskip=3pt}
    \caption{
    \textbf{Success rate under different threshold.}
    }
    \vspace{-0.7cm}
    \label{fig:sr_threhold}
\end{wrapfigure}

We present a comprehensive breakdown of the success rates in real-world long-horizon settings under different thresholds in \figref{fig:sr_threhold}, where our method consistently outperforms all baselines.

\begin{figure}[t!]
    \centering
    \includegraphics[width=1\textwidth]{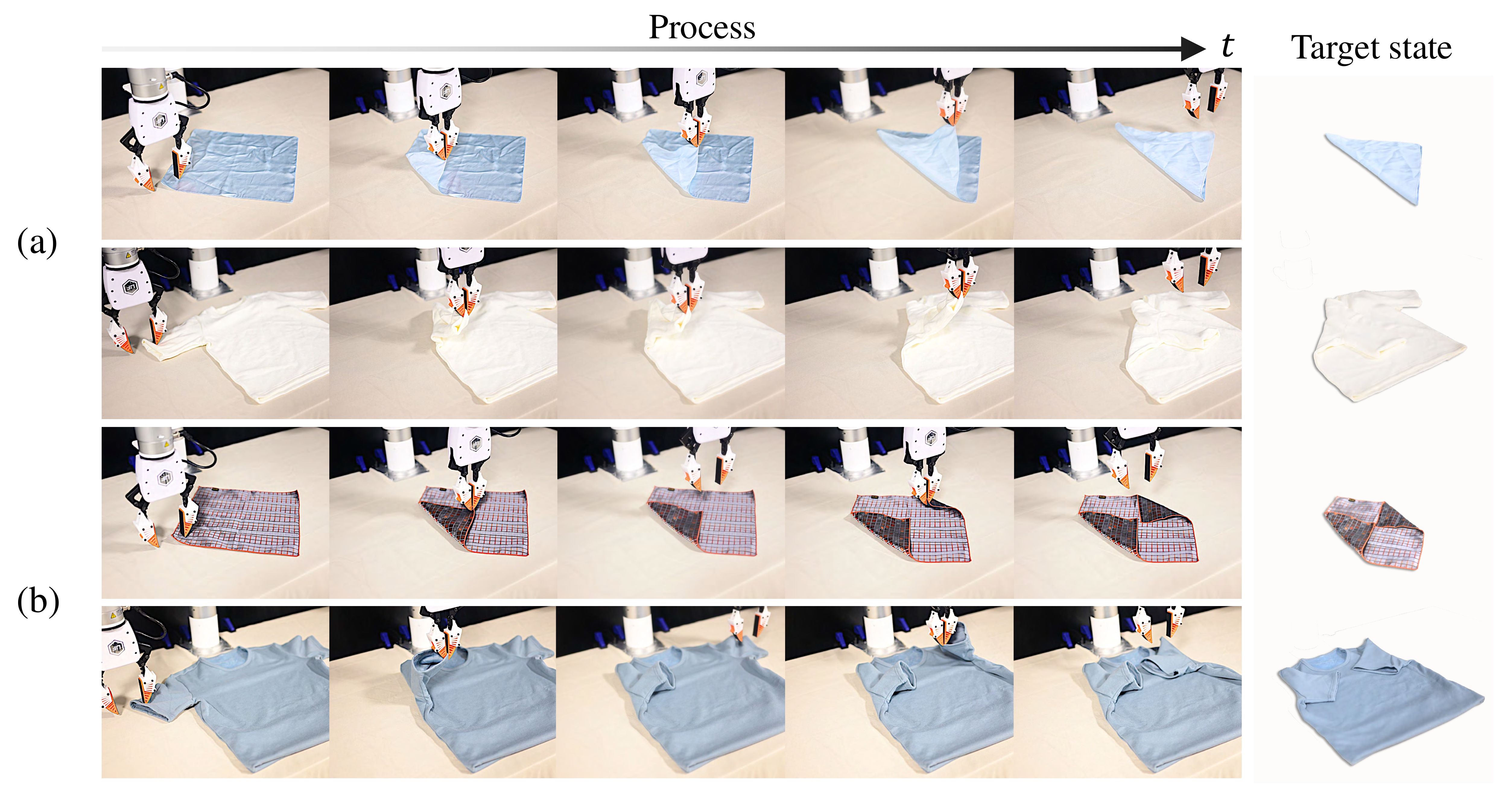}
    \caption{\textbf{Qualitative results of real-world system deployment.} The target state is represented in the last column of each row. Each experimental sequence illustrates the progressive deformation states during folding tasks. The first two rows correspond to single-step scenarios, while the last two represent multi-step scenarios.}
    \label{fig:planning_qualitative_results}
\end{figure}

\paragraph{Additional Qualitative Results}
Accordingly, we also provide additional qualitative results for single-step and multi-step scenarios on clothes and T-shirts in \figref{fig:planning_qualitative_results}. The results validate our system's capability to accurately manipulate diverse fabric items from arbitrary initial configurations to challenging target folding states.

\begin{figure}[h]
\centering
\includegraphics[width=0.8\textwidth]{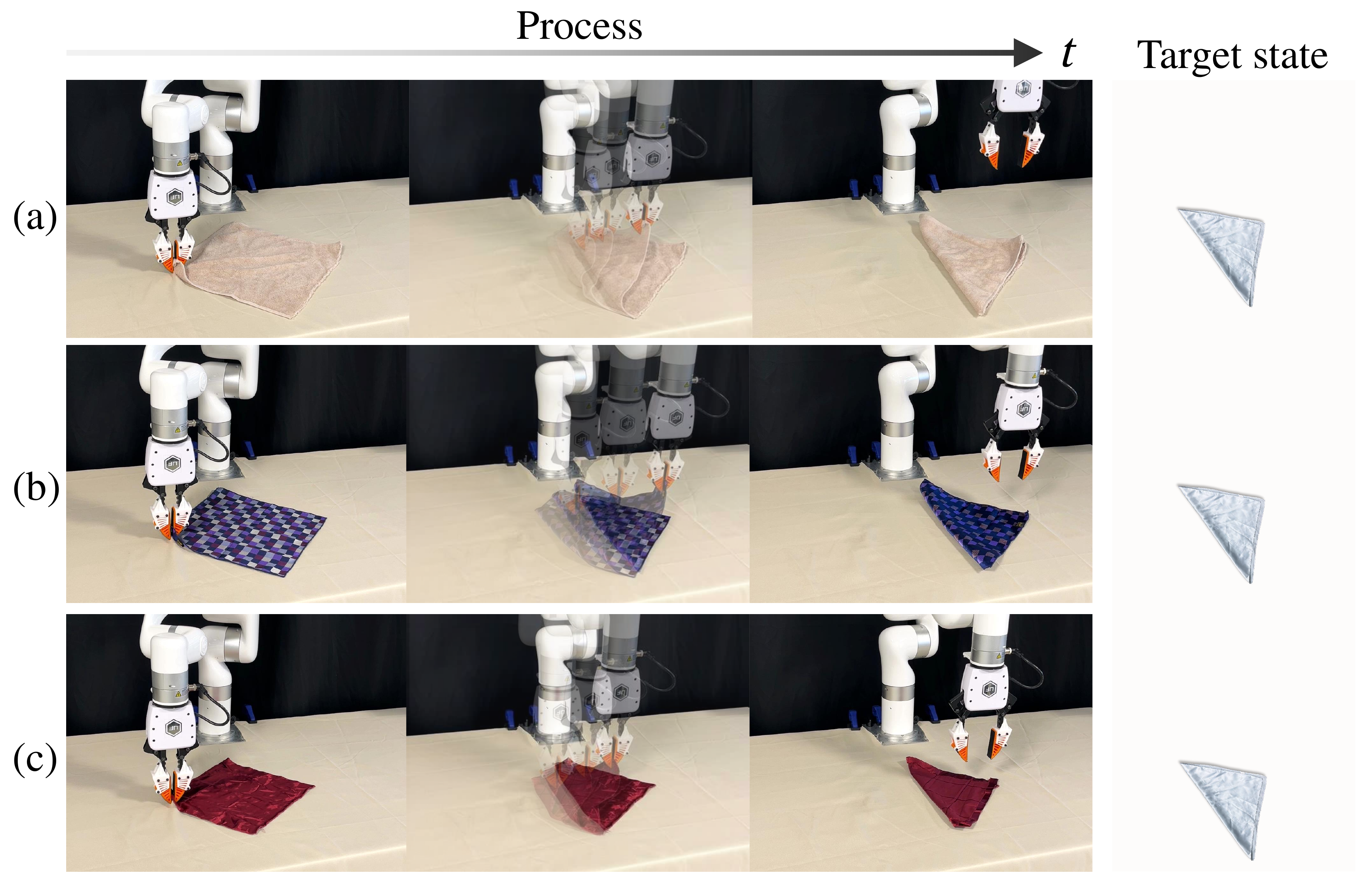}
\caption{\textbf{Intra-class generalization evaluation.} We demonstrate that our method can generalize across garments with varying physical attributes (size, material, and color). The garment size progressively decreases from (a) to (c).}
\label{fig:intra_class}
\end{figure}

\paragraph{Additional Intra-class Generalization Results}
We provide additional qualitative results for intra-class generalization on square cloth object in \figref{fig:intra_class} with sizes ranging from 20 cm to 40 cm. Our model successfully executes precise folding trajectories across these variations, consistently achieving target configurations and confirming robust intra-class generalization.


\clearpage

\subsection{Simulation Planning Results}
\begin{figure*}[h]
    \centering
    \begin{subfigure}{0.24\textwidth}
    \centering
    \includegraphics[width=\textwidth]{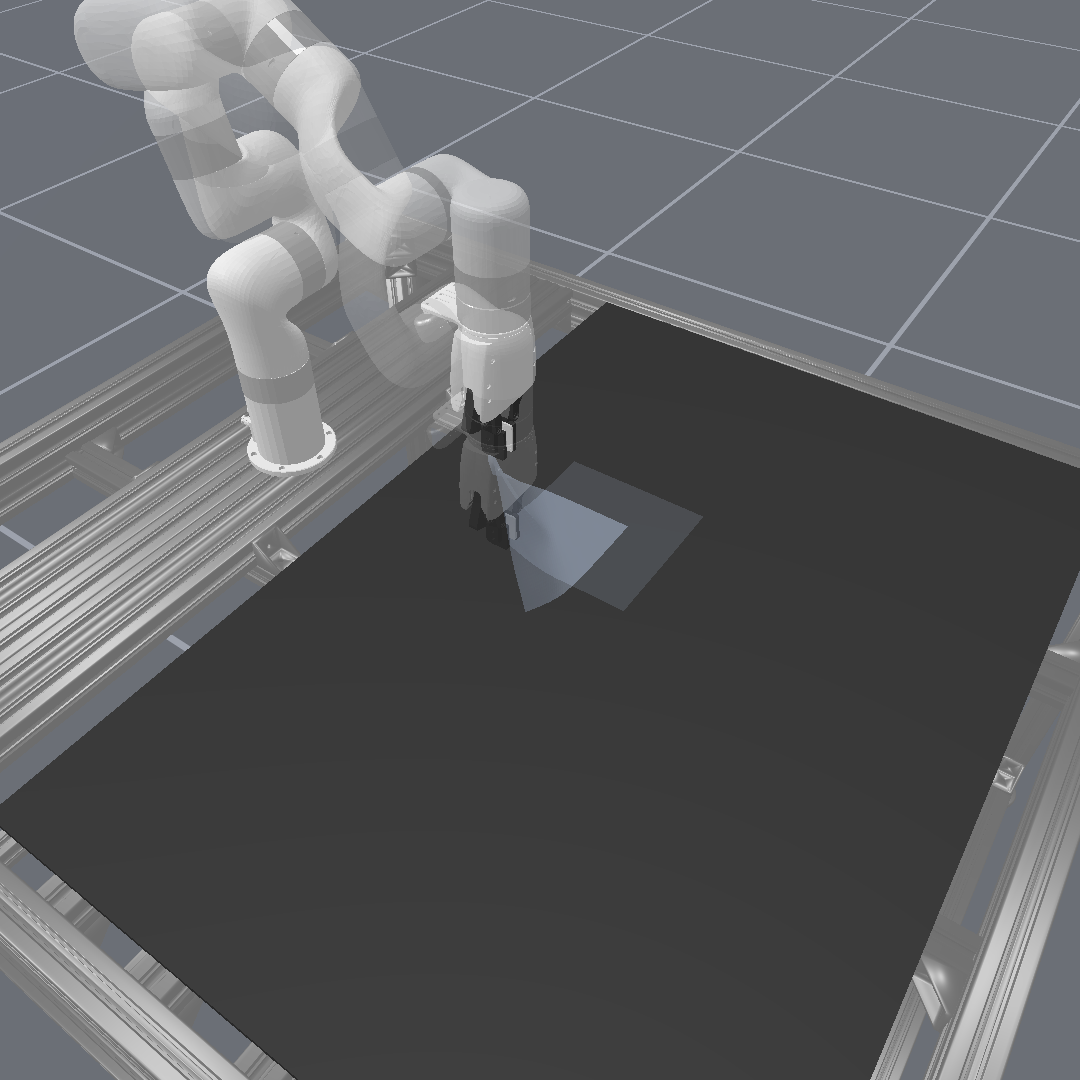}
    \caption{Lift cloth}
    \label{fig:sub1}
    \end{subfigure}
    \hfill
    \begin{subfigure}{0.24\textwidth}
    \centering
    \includegraphics[width=\textwidth]{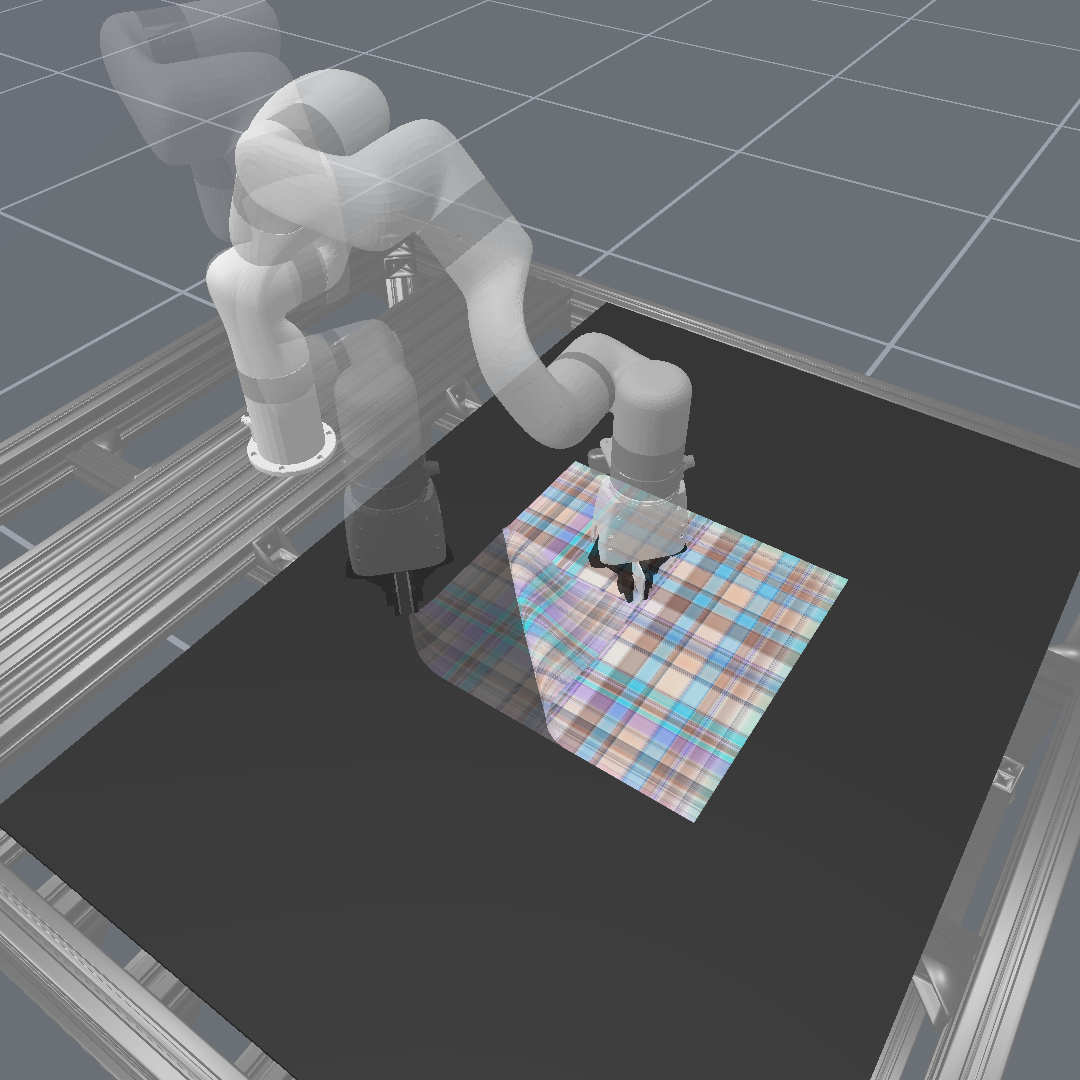}
    \caption{Fold cloth}
    \label{fig:sub2}
    \end{subfigure}
    \hfill
    \begin{subfigure}{0.24\textwidth}
    \centering
    \includegraphics[width=\textwidth]{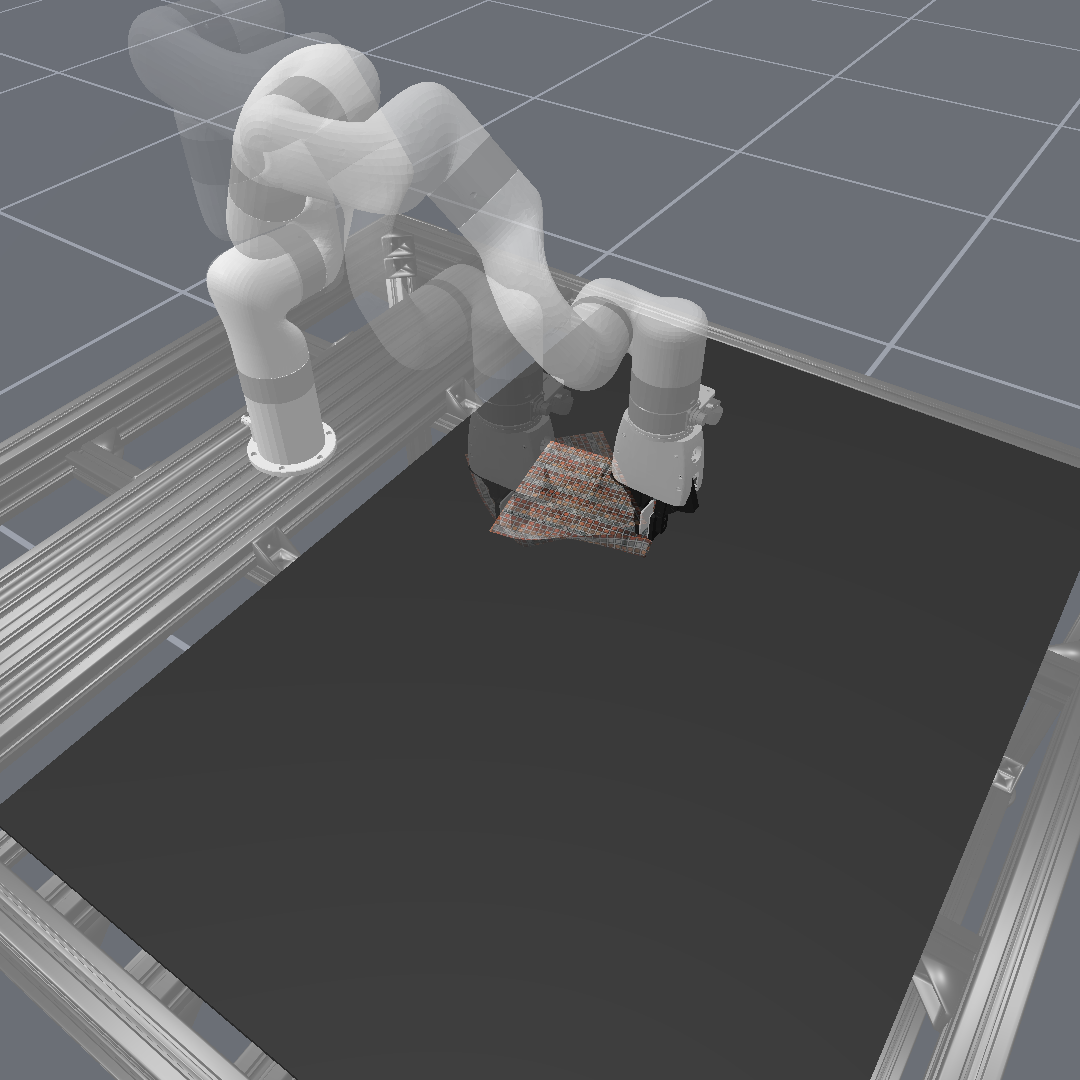}
    \caption{Rotate cloth}
    \label{fig:sub3}
    \end{subfigure}
    \hfill
    \begin{subfigure}{0.24\textwidth}
    \centering
    \includegraphics[width=\textwidth]{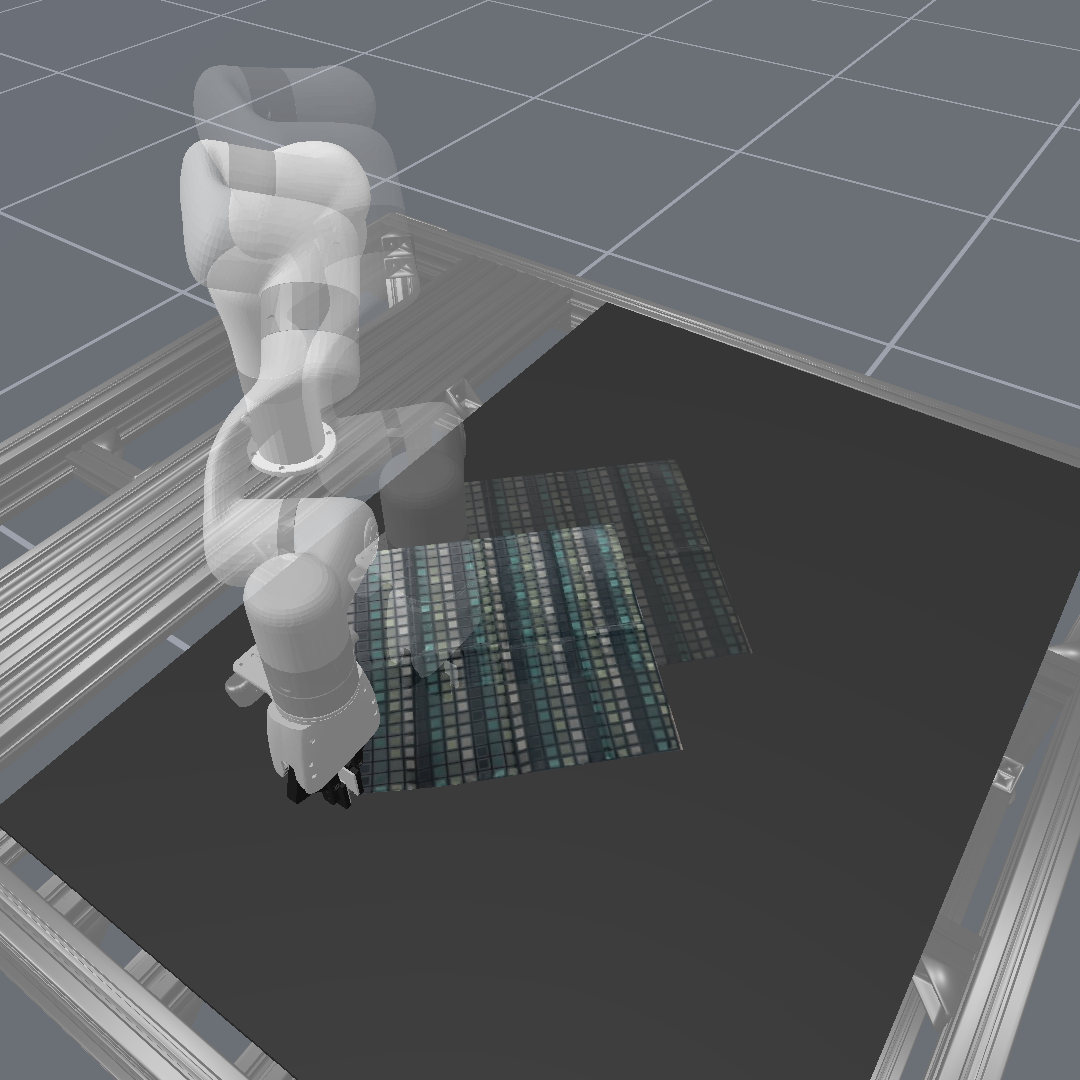} 
    \caption{Pull cloth}
    \label{fig:sub4}
    \end{subfigure}
\caption{\textbf{Simulated cloth manipulation environments.} Visualization of diverse manipulation scenarios in simulation: (a)-(d) demonstrate different cloth-robot interactions with varied object configurations and manipulation tasks.}
\label{fig:simenv}
\vspace{-10pt}  
\end{figure*}

\begin{figure*}[h]
\begin{center}
\includegraphics[width=0.95\linewidth]{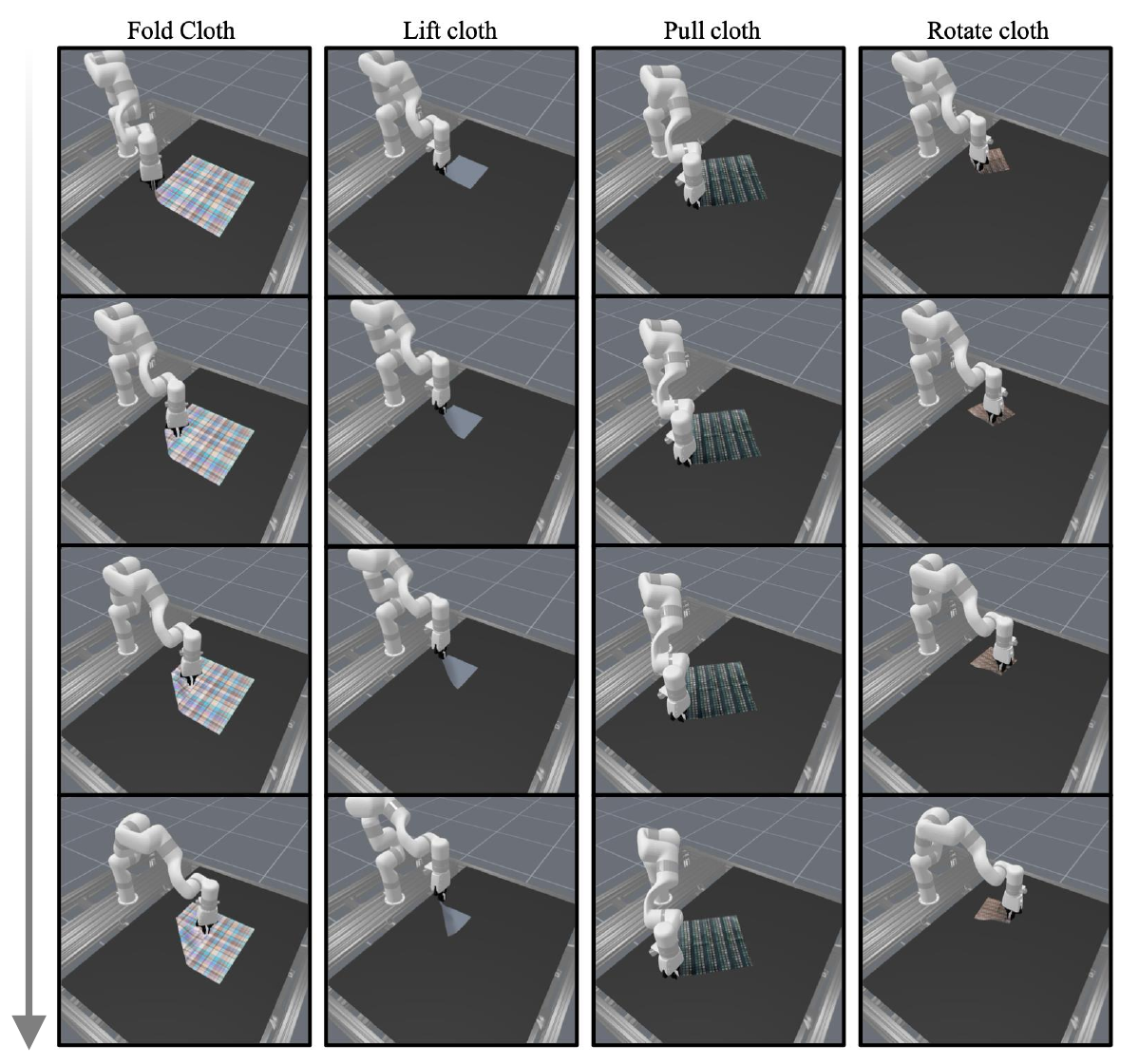}
\end{center}
\caption{\textbf{Model predictive control evaluation in simulation.} Demonstration of our diffusion-based dynamics model integrated with MPC across diverse manipulation tasks using xArm7, validated on various cloth types.}
\label{fig:planning_qulitative_sim}
\end{figure*}

We present more qualitative results in \figref{fig:planning_qulitative_sim} in the simulation environment on planning. We design four simple tasks in simulation for system validation as visualized in \figref{fig:simenv}.

\clearpage
\section{Experiment Setup}\label{sec:experiment_setup}
\subsection{Task Description}\label{sec::tast_description}

\begin{wrapfigure}{r}{0.5\textwidth}
\vspace{-1cm}
  \centering
  \footnotesize
  \setlength\tabcolsep{3pt}
  \renewcommand{\arraystretch}{1.05}

  \includegraphics[width=0.5\textwidth]{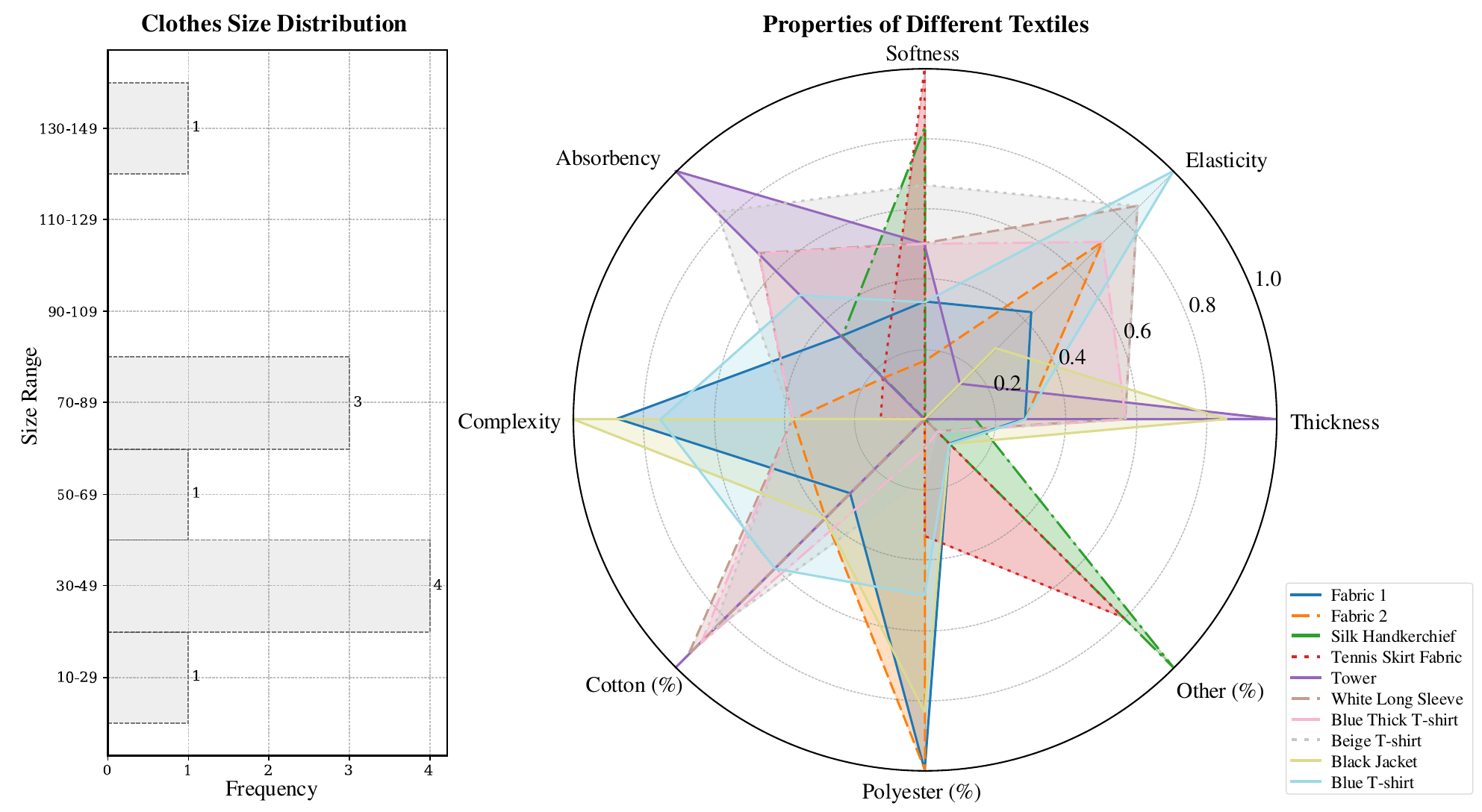}
  \caption{\textbf{Coverage of different clothes in our experiments.}} 
  \label{fig:material}

   \includegraphics[width=0.5\textwidth]{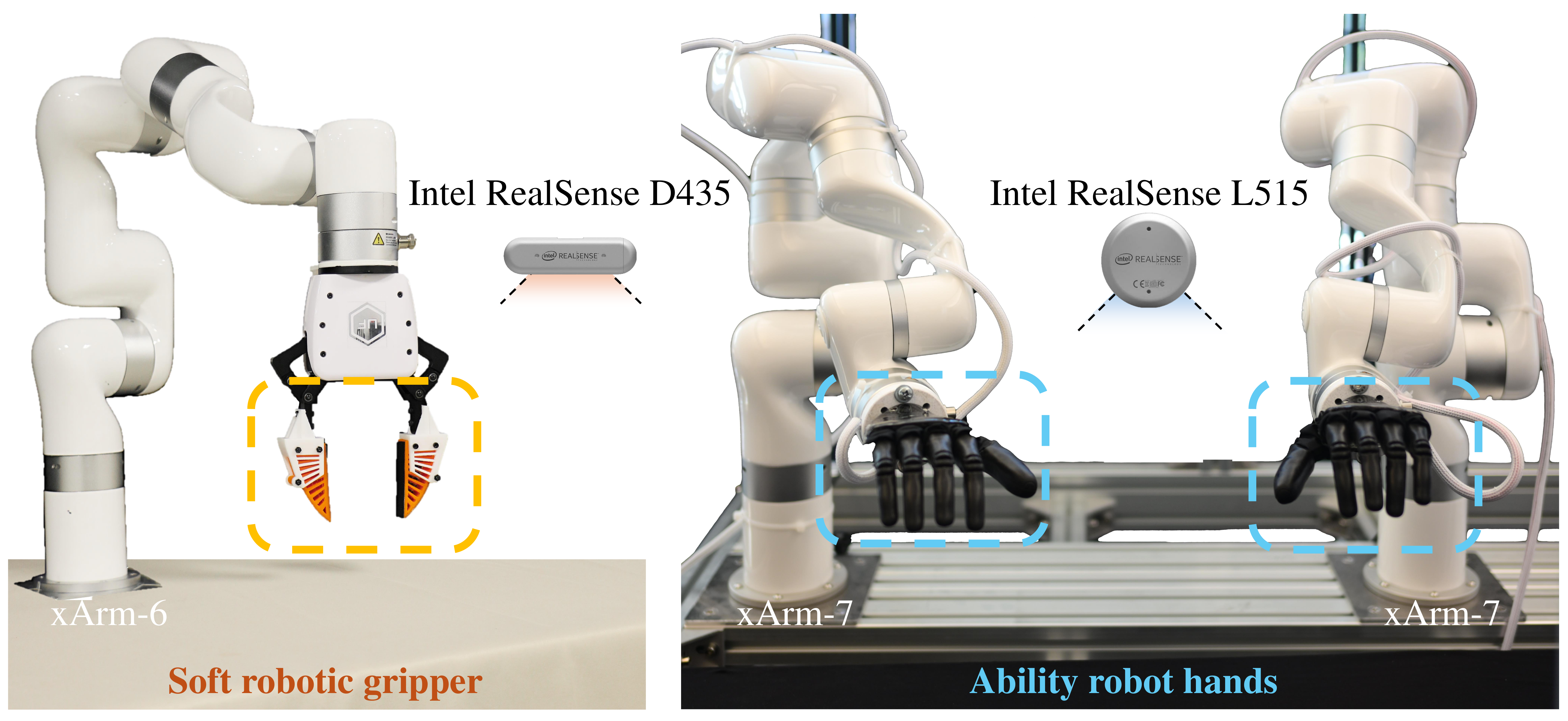}
    \caption{\textbf{Hardware overview.} Our real-world platform includes a UFactory xArm-6 and a bimanual dexterous system consisting of two UFactory xArm-7 robots with Ability hands. Each robot is equipped with one RGB-D camera.
    }
    \label{fig:robot_hardware_steup}

    \includegraphics[width=0.5\textwidth]{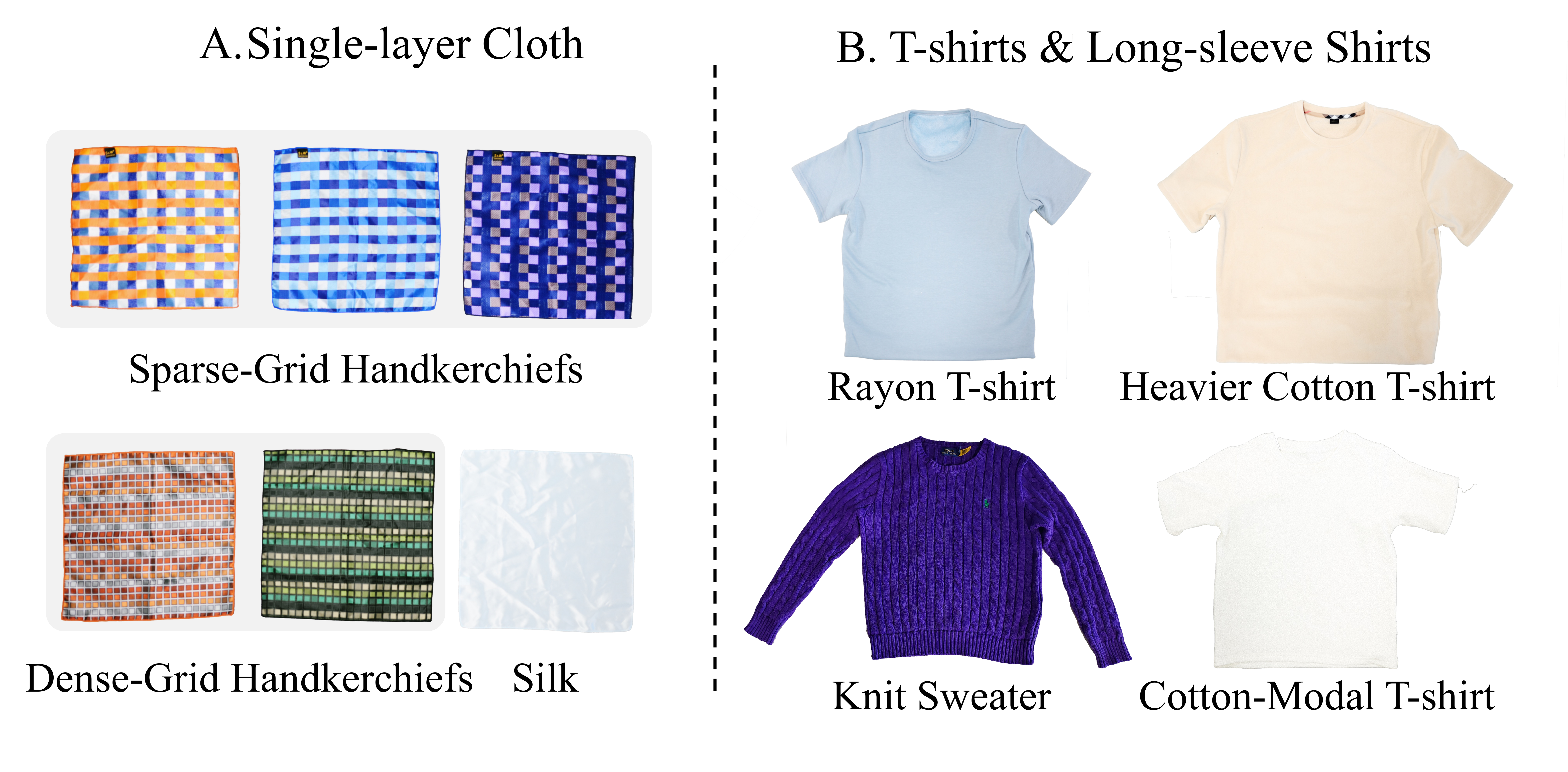}
    \caption{\textbf{Cloth overview.} We evaluate our method on different single-layer clothes as well as dual-layer T-shirts and long-sleeve shirts with varying colors and materials.}
    \vspace{-0.5cm}
    \label{fig:cloth_assets}

\end{wrapfigure}

We evaluate our method on challenging cloth folding tasks characterized by significant visual occlusion and complex physical dynamics, demonstrating the real-world performance of our diffusion-based perception and dynamics model.

\textbf{Square Cloth folding and unfolding.} This task explores robotic cloth folding and unfolding tasks across diverse fabrics. We employ prediction results from DPM to define target shapes, enabling accurate shape matching between the manipulated cloth and desired folding configurations. The system aims to robustly handle variations in fabric characteristics while maintaining folding accuracy. This task is more challenging than usual pushing or relocating tasks due to significant visual occlusions during the folding process, and the increased action complexity. Achieving precise folding to a specified target configuration requires both an accurate estimation and the dynamic prediction of the cloth. We tested with square handkerchiefs made of three different materials. Each of these clothes has a different visual appearance and size. 
        
\textbf{Garment folding and unfolding.} This task focuses on folding or unfolding a T-shirt or a long-sleeve top into the target configuration. Such garments present unique challenges due to their dual-layer structure and compliant dynamics. We evaluate our approach on garments of different sizes and physical properties. We set more challenging target states (such as diagonal fold and fold in half) that require higher motion accuracy. Incorrect actions will increase the recovery cost. Some target states also require changing the grasp contact points and performing multiple folds. \figref{fig:cloth_assets} shows all the test cloths and garments with various materials and sizes used in our real-world experiments.

We visualize the distribution of all the clothes and garments in our experiments in \figref{fig:material}

\subsection{Physical Setup}\label{appendix:physical_setup}
We validate our system on two robotic platforms: (1) a single UFactory xArm-6 robotic arm with Fin Ray Effect-based soft robotic fingers for gripping cloth, and (2) a stationary bimanual dexterous system consisting of two UFactory xArm-7 robotic arms, each equipped with a 6-DoF Ability hand. Both setups use a single RGB-D camera: the Intel RealSense D435 with 640 $\times$ 480 resolution for the xArm-6 and the L515 with 1024 $\times$ 768 resolution for the dual-arm system. \figref{fig:robot_hardware_steup} illustrates our hardware setup.

\section{Implementation Details}

\subsection{Data Collection}\label{sec::data_collection}
We collect training data for learning state estimation and dynamics prediction in a simulation environment built on SAPIEN~\citep{SAPIEN}. The rigid bodies, such as the robot arm, are simulated using the built-in PhysX-based simulator, while the cloth is simulated with the projective dynamics (PD) solver \citep{Bouaziz2014ProjectiveDynamics}. The two systems are coupled at the time step level by alternating updates: the PD system treats the positions and velocities of PhysX-managed objects as boundary conditions, and PhysX does the same for the PD-managed cloth. In the PD system, the cloth is modeled as a hyper-elastic thin shell. We follow \citet{Ly2020DryFrictionalPD} to simulate collision and friction in the PD system. We provide detailed physical parameters for cloth simulation in \tabref{tab:sim_params}.

\begin{wrapfigure}{r}{0.3\textwidth}
  \centering
  \footnotesize
  \setlength\tabcolsep{3pt}
  \renewcommand{\arraystretch}{1.05}

  \begin{tabular}{@{}lp{10mm}}
    \toprule
    Physical Parameter & Value\\
    \midrule
    collision margin & 1e-3 \\
    collision weight & 5e3 \\
    collision sphere radius & 8e-3 \\
    damping & 1e-2 \\
    thickness & 1e-3 \\
    density & 1e3 \\
    stretch stiffness & 1e3 \\
    bend stiffness & 1e-3 \\
    friction & 0.5 \\
    gravity & -9.81 \\
    \bottomrule
  \end{tabular}
  \captionof{table}{Simulation physical parameters.}
  \vspace{-0.3cm}
  \label{tab:sim_params}
\end{wrapfigure}

To collect state estimation data, we set up a comprehensive multi-view system that incorporates up to four calibrated stereo-depth sensors, strategically placed at randomized viewing angles within predefined ranges. Cloth is initialized given a randomly "pick-and-place" action. This configuration enables the generation of paired datasets consisting of fused point clouds alongside their corresponding ground-truth mesh states across multiple viewpoints. The system leverages SAPIEN's advanced stereo depth simulation capabilities\citep{10027470}, which significantly reduces the sim-to-real gap by faithfully reproducing point cloud characteristics observed in real-world scenarios. This high-fidelity simulation approach ensures robust and reliable state estimation performance when transferred to physical environments. In our point cloud fusion process, we augment camera extrinsic parameters to simulate real-world calibration errors. Specifically, we introduce rotational variations ranging from $-1.5^{\circ}$ to $1.5^{\circ}$ and translational variations from -0.5 to 0.5 cm. To better mimic real-world conditions, we also simulate depth sensor noise and occlusion effects by applying random point dropout with ratios between 0.1 and 0.2, and introducing noise to the fused point cloud.

To collect dynamic data, we employ diverse action sampling strategies to generate a comprehensive dataset of 500K examples. Our sampling approach encompasses two key methodologies designed to capture realistic cloth manipulation scenarios. The first method involves applying directionally-randomized displacements to selected mesh vertices, with particular emphasis on folding-oriented actions where the cloth is manipulated to create various folding patterns. We also simulate picking and relocation actions by applying upward and translational movements to randomly selected vertices. The second methodology focuses on pair-wise vertex manipulation, where vertex pairs are selected based on their spatial distances to simulate actions such as folding one point of the cloth onto another. Each incremental action is precisely controlled, with magnitudes ranging from 0.02 to 0.05 units. To evaluate the model's performance across different time horizons and assess the impact of auto-regressive inference error accumulation, we generate action sequences varying in length from 15 to 35 steps. All resultant mesh deformations throughout these sequences are meticulously recorded to capture the complete dynamics of the cloth's behavior.

\subsection{Model Details}
\label{appendix:model_details}
\paragraph{Point Cloud Encoder}
We employ a patch-based architecture for point cloud encoding that processes the input through local grouping and feature extraction. The encoder first groups points using a KNN-based strategy, then processes each local patch through a specialized patch encoder, and finally incorporates positional information through learnable embeddings. This design enables effective capture of both local geometric structures and global spatial relationships.

\begin{table}[h]
\centering
\begin{tabular}{ll}
\hline
Hyperparameter & Value \\
\hline
Output dimension & 1024 \\
Number of groups & 256 \\
Group size & 64 \\
Group radius & 0.15 \\
Position embedding dimension & 128 \\
Patch encoder hidden dims & [128, 512] \\
\hline
\end{tabular}
\vspace{0.2cm}
\caption{Point cloud encoder hyperparameters.}
\vspace{-0.5cm}
\end{table}

\paragraph{Model Architecture}
We design a transformer-based architecture for state estimation, which consists of a point cloud encoder, a positional embedding module, and a series of transformer blocks. The model takes both point cloud observations and mesh states as input. In dynamics model, the model takes an additional input channel containing
a binary mask, indicating grasped mesh vertices. The point cloud is first processed through a patch-based encoder, while the mesh states are embedded using a patchified positional encoding scheme. These features are then processed through transformer blocks with cross-attention mechanisms to predict the mesh state.

\begin{table}[h]
\centering
\begin{tabular}{ll}
\hline
Hyperparameter & Value \\
\hline
Number of attention heads & 16 \\
Attention head dimension & 88 \\
Number of transformer layers & 4 \\
Inner dimension & 1408 \\
Dropout & 0.0 \\
Cross attention dimension & 1024 \\
Point cloud embedding dimension & 1024 \\
Number of input frames & 2 \\
Number of output frames & 1 \\
Activation function & GELU \\
Output MLP dimensions & [512, 256] \\
Normalization type & AdaLayerNorm \\
Normalization epsilon & 1e-5 \\
\hline
\end{tabular}
\vspace{0.2cm}
\caption{Model hyperparameters.}
\vspace{-0.5cm}
\end{table}

\begin{figure*}
    \centering
    \includegraphics[width=1\linewidth]{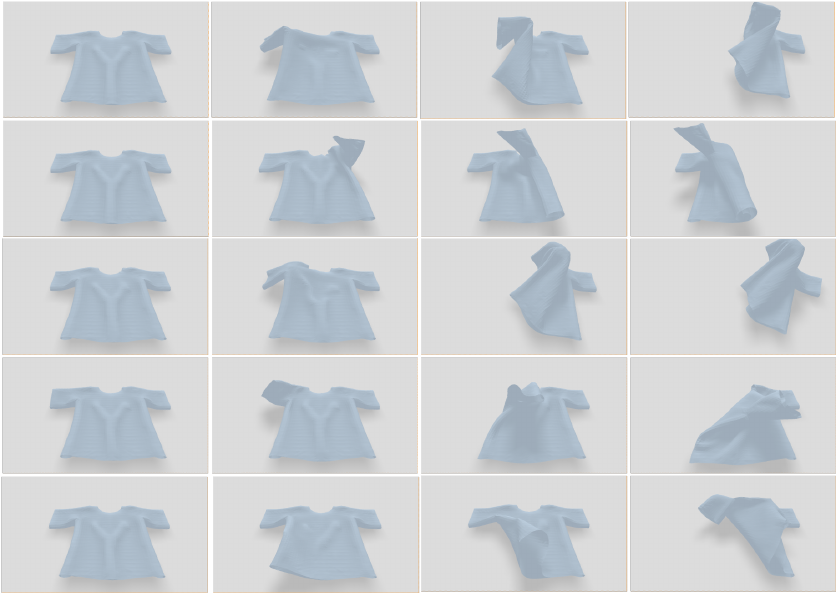}
    \caption{\textbf{Example training data.}}
    \label{fig:enter-label}
\end{figure*}

\paragraph{Action Embedding}
We employ a Fourier feature-based action encoding scheme to effectively represent mesh manipulation actions in a high-dimensional space. The action encoder consists of two main components: (1) a Fourier feature mapping that projects 3D action vectors into a higher-dimensional space using sinusoidal functions, and (2) a multi-layer perceptron that further transforms these features into the desired embedding dimension.

\begin{wrapfigure}{r}{0.45\textwidth}
  \centering
  \footnotesize
  \setlength\tabcolsep{6pt}
  \renewcommand{\arraystretch}{1.1}
    \vspace{-0.5cm}
  \begin{tabular}{ll}
    \toprule
    Hyperparameter & Value \\
    \midrule
    Fourier frequencies & 8 \\
    Fourier feature dimension & 48 \\
    MLP hidden dimensions & [512, 512] \\
    Output dimension & output\_dim \\
    Activation function & SiLU \\
    Position normalization & Center \& Scale \\
    \bottomrule
  \end{tabular}
  \captionof{table}{Action encoder hyperparameters.}
  \label{tab:action_encoder_hyperparams}
\end{wrapfigure}

The Fourier feature mapping applies frequency-based encoding separately to each spatial dimension $(A_x, A_y, A_z)$ of the action vectors using both sine and cosine functions, resulting in an intermediate representation of dimension $2 \times 3 \times F$, where $F$ is the number of Fourier frequencies.  Given the input action $a \in \mathbb{R}^{B\times N\times3}$, where $N$ is the number of actions and $3$ represents the dimension of $(x, y, z)$ coordinates, we compute the embedding $\mathbf{e} \in \mathbb{R}^{B \times N \times D_3}$ as:
\begin{equation}
\mathbf{e}_{b,n,d} = 
\begin{bmatrix}
\sin(2\pi f_d a_{b,n,x}) & \cos(2\pi f_d a_{b,n,x}) \\
\sin(2\pi f_d a_{b,n,y}) & \cos(2\pi f_d a_{b,n,y}) \\
\sin(2\pi f_d a_{b,n,z}) & \cos(2\pi f_d a_{b,n,z})
\end{bmatrix}
\end{equation}
where $D_3$ is the action embedding dimension, $d \in \{0, \ldots, D_3/6 - 1\}$, and $f_d = 100^{d / (D_3/6)}$ are the Fourier feature frequencies. The resulting embedding $e$ provides a rich, high-dimensional representation of the action space. This representation is then processed through an MLP to produce the final action embeddings, which is later injected as the condition into our model through cross-attention layers.

\subsection{Training Details}
\label{appendix:training_details}
We train our model using distributed data parallel training on 4 H100 GPUs. The model is trained with a batch size of 128 per GPU and gradient accumulation steps of 4, resulting in an effective batch size of 2048. We use the AdamW optimizer with a learning rate of 1e-5 and cosine learning rate scheduler with 1000 warmup steps. For numerical stability and training efficiency, we employ mixed-precision training with bfloat16 and enable TF32 on supported hardware.

\begin{wrapfigure}{r}{0.4\textwidth}
  \centering
  \footnotesize
  \setlength\tabcolsep{8pt}
  \renewcommand{\arraystretch}{1.1}

  \begin{tabular}{ll}
    \toprule
    Hyperparameter & Value \\
    \midrule
    Number of GPUs & 4 \\
    Batch size per GPU & 128 \\
    Gradient accumulation steps & 4 \\
    Effective batch size & 1024 \\
    Learning rate & 1e-5 \\
    Learning rate scheduler & Cosine \\
    Warmup steps & 1000 \\
    Mixed precision & bfloat16 \\
    Number of workers & 16 \\
    \bottomrule
  \end{tabular}
  \captionof{table}{Training hyperparameters.}
  \label{tab:training_hyperparams}
  \vspace{-0.5cm}
\end{wrapfigure}

\subsection{Planning Details}
\label{appendix:planning_details}
For planning, we employ a hybrid approach combining Model Predictive Control (MPC) and Cross Entropy Method (CEM). Our planner optimizes action sequences by iteratively sampling actions, evaluating their outcomes using the learned dynamics model, and updating the sampling distribution based on the costs.
To enhance planning efficiency, we introduce two key strategies: (1) an informed action sampling mechanism and (2) a grasp point selection method. For action sampling, we initialize the sampling distribution using a prior direction informed by the target state. Specifically, we identify the K vertices with the highest mean squared error (MSE) between the current and target states, and compute a weighted average direction based on their distances to the grasp point:

\begin{equation}
    d_{main} = \sum_{i=1}^K w_i(s_t^i - s_c^i), \quad w_i = \frac{1}{\|p_g - p_i\| + \epsilon}
\end{equation}

where $s_t^i$ and $s_c^i$ are target and current states of vertex $i$, $p_g$ is the grasp point position, and $p_i$ is the position of vertex $i$. This informed direction guides the initial sampling distribution for more efficient exploration.

For grasp point selection, we employ a temperature-controlled softmax strategy based on vertex displacements. Given the current state $S_c$ and target state $S_t$, we compute a probability distribution over all vertices:

\begin{equation}
    p(i) = \frac{\exp(\|s_t^i - s_c^i\|_2/\tau)}{\sum_j \exp(\|s_t^j - s_c^j\|_2/\tau)}
\end{equation}

where $s_t^i$ and $s_c^i$ represent the position of vertex $i$ in target and current states, respectively, and $\tau$ is a temperature parameter that controls the concentration of the probability distribution. A lower temperature leads to more deterministic selection focusing on maximum displacement vertices, while a higher temperature enables more exploratory behavior. The grasp point is then sampled from this distribution:

\begin{equation}
    g \sim p(i)
\end{equation}

This probabilistic selection mechanism provides several advantages over deterministic maximum displacement selection: (1) it allows for exploration of different grasp points, (2) it can adapt to different manipulation scenarios by adjusting the temperature parameter, and (3) it provides a smoother transition between different grasp point candidates.
The planning algorithm is outlined in Algorithm \ref{alg:planning}.
Hyperparameters for model-based planning are listed in \tabref{tab:planning_param}.

\begin{wrapfigure}{r}{0.48\textwidth}
  \centering
  \footnotesize
  \setlength\tabcolsep{6pt}
  \renewcommand{\arraystretch}{1.1}

  \begin{tabular}{ll}
    \toprule
    Parameter & Value \\
    \midrule
    Number of iterations & 5 \\
    Samples per iteration & 16 \\
    Sequence length & 5 \\
    Action dimension & 3 \\
    Initial std deviation & 0.1 \\
    Temperature & 1.0 \\
    \bottomrule
  \end{tabular}
  \captionof{table}{Planning hyperparameters.}
  \label{tab:planning_param}
  \vspace{0.5cm}

  \begin{tabular}{ll}
    \toprule
    Hyperparameter & Value \\
    \midrule
    Maximum particles ($N_{obj}$) & 100 \\
    Maximum relations ($N_R$) & 1000 \\
    History frames ($n_{his}$) & 3 \\
    Future frames ($n_{future}$) & 5 \\
    State dimension & 3 \\
    Attribute dimension & 2 \\
    FPS radius range & [0.05, 0.1] \\
    Adjacency radius range & [0.74, 0.76] \\
    Topk neighbors & 5 \\
    \bottomrule
  \end{tabular}
  \captionof{table}{GNN model hyperparameters.}
  \label{tab:gnn_hyperparams}
  \vspace{-1cm}

\end{wrapfigure}
\subsection{State Estimation Baseline Implementation}
To create the fairest possible comparison, we provided the GT canonical mesh to both GarmentNets and MEDOR. This isolates the evaluation to their performance for mapping a known shape to a deformed configuration in the observation space. We retrained the TRTM baseline from scratch on our data, and evaluated GarmentNets and MEDOR using their official pretrained checkpoints. The model input domain gap is minimal, as both our work and these baselines use the CLOTH3D dataset~\citep{bertiche2020cloth3d} with the same crumpled-state generation procedures. The evaluation is particularly fair for MEDOR for its test-time adaptation mechanism.

\subsection{Dynamics Baseline Implementation}
We introduce details of the dynamics baseline implementation.

\paragraph{GNNs} We adopt the implementation from \citep{zhang2024adaptigraph}.
We construct a comprehensive graph representation for modeling cloth dynamics, incorporating object particles, end-effector interactions, and material properties. The graph structure consists of four main components: (1) state and action representations, (2) particle attributes and instance information, (3) relation matrices for particle interactions, and (4) material-specific physics parameters.
The state representation captures both spatial positions and temporal dynamics through a history buffer of $n_{his}$ frames and future predictions of $n_{future}$ frames. Each state vector contains the 3D positions $(x, y, z)$ of both cloth particles and the end effector. We maintain a fixed-size particle set through Farthest Point Sampling with an adaptive radius range of $[0.05, 0.1]$.  We show detailed parameters for graph construction in \tabref{tab:gnn_hyperparams}.

\subsection{Manipulation Pipeline Details}
Our system integrates OWLV2~\citep{minderer2024scaling} and Segment Anything~\citep{kirillov2023segany} to detect and segment desktop objects from RGB-D input. A single-view partial point cloud of the target object serves as input, which is processed via DPM to infer the state of the cloth. To address the dimensional and positional discrepancies between predicted and observed point clouds, we implement a two-stage alignment process. First, we compute the spatial dimensions of the observed point cloud and apply appropriate scaling transformations to the predicted point cloud. Subsequently, we employ the Iterative Closest Point (ICP) algorithm for fine-grained alignment, ensuring that MPC-generated grasping positions and motion trajectories can be accurately mapped to the physical object. For manipulation, we model both soft robotic grippers and dexterous hands by representing their end effectors as particles that attach to mesh vertices during motion. To evaluate our system, we first collect realistic and challenging target states through teleoperation. We then conduct 10 experimental trials for the same target state, executing a delta action sequence through the MPC with the dynamics model. These actions are transformed into absolute positions in the base frame of the robotic arm, with smooth Cartesian trajectories generated using joint online trajectory planning.

\begin{algorithm*}
\caption{MPC Planning Algorithm}
\label{alg:planning}
\begin{algorithmic}[1]
\Require Initial state $s_i$, target state $s_t$, dynamics model $f_\theta$, number of iterations $N$
\Require Number of samples $K$, sequence length $L$, action bounds $[a_{min}, a_{max}]$
\State Initialize $\mu \gets \mathbf{0}$, $\sigma \gets 0.1$
\State $a_{best} \gets \text{None}$, $c_{best} \gets \infty$
\For{$i = 1$ to $N$}
    \State $A_{mppi} \gets$ SampleGaussian$(K/2, L, \mu, \sigma, [a_{min}, a_{max}])$
    \State $A_{uniform} \gets$ SampleUniform$(K/2, L, [a_{min}, a_{max}])$
    \State $A \gets$ Concatenate$(A_{mppi}, A_{uniform})$
    \State $S_{pred} \gets f_\theta(S, A)$ \Comment{Predict trajectories}
    \State $C \gets$ ComputeCost$(S_{pred}, A, T)$ \Comment{Evaluate costs}
    \If{$\min(C) < c_{best}$}
        \State $c_{best} \gets \min(C)$
        \State $a_{best} \gets A[\arg\min(C)]$
    \EndIf
    \State $\mu, \sigma \gets$ UpdateDistribution$(A, C, \tau)$ \Comment{Update using weighted averaging}
    \State $\sigma \gets \sigma \cdot (1 - i/N)$ \Comment{Anneal exploration}
\EndFor
\State \Return $a_{best}$
\end{algorithmic}
\label{alg:planning}
\end{algorithm*}

%% file: main.bbl
\begin{thebibliography}{51}
\providecommand{\natexlab}[1]{#1}
\providecommand{\url}[1]{\texttt{#1}}
\expandafter\ifx\csname urlstyle\endcsname\relax
  \providecommand{\doi}[1]{doi: #1}\else
  \providecommand{\doi}{doi: \begingroup \urlstyle{rm}\Url}\fi

\bibitem[Longhini et~al.(2024-12-02)Longhini, Wang, Garcia-Camacho, Blanco-Mulero, Moletta, Welle, Alenyà, Yin, Erickson, Held, Borràs, and Kragic]{Longhini2024unfolding}
A.~Longhini, Y.~Wang, I.~Garcia-Camacho, D.~Blanco-Mulero, M.~Moletta, M.~Welle, G.~Alenyà, H.~Yin, Z.~Erickson, D.~Held, J.~Borràs, and D.~Kragic.
\newblock Unfolding the literature: A review of robotic cloth manipulation.
\newblock \emph{Annual Review of Control, Robotics, and Autonomous Systems}, 2024-12-02.
\newblock ISSN 2573-5144.

\bibitem[Yin et~al.(2021)Yin, Varava, and Kragic]{yin2021modeling}
H.~Yin, A.~Varava, and D.~Kragic.
\newblock Modeling, learning, perception, and control methods for deformable object manipulation.
\newblock \emph{Science Robotics}, 6\penalty0 (54):\penalty0 eabd8803, 2021.

\bibitem[Chi and Song(2021)]{chi2021garmentnets}
C.~Chi and S.~Song.
\newblock Garmentnets: Category-level pose estimation for garments via canonical space shape completion.
\newblock In \emph{The IEEE International Conference on Computer Vision (ICCV)}, 2021.

\bibitem[Huang et~al.(2022)Huang, Lin, and Held]{Huang2022MeshbasedDW}
Z.~Huang, X.~Lin, and D.~Held.
\newblock Mesh-based dynamics with occlusion reasoning for cloth manipulation.
\newblock \emph{ArXiv}, abs/2206.02881, 2022.
\newblock URL \url{https://api.semanticscholar.org/CorpusID:248942073}.

\bibitem[wan(2024)]{wang2023trtm}
Trtm: Template-based reconstruction and target-oriented manipulation of crumpled cloths.
\newblock 2024.

\bibitem[Zhang et~al.(2024)Zhang, Li, Hauser, and Li]{zhang2024adaptigraph}
K.~Zhang, B.~Li, K.~Hauser, and Y.~Li.
\newblock Adaptigraph: Material-adaptive graph-based neural dynamics for robotic manipulation.
\newblock In \emph{Proceedings of Robotics: Science and Systems (RSS)}, 2024.

\bibitem[Li et~al.(2018)]{li2018learning}
Y.~Li et~al.
\newblock Learning particle dynamics for manipulating rigid bodies, deformable objects, and fluids.
\newblock \emph{arXiv preprint arXiv:1810.01566}, 2018.

\bibitem[He et~al.(2025)He, Ai, Liu, Wan, Christensen, and Su]{he2025learning}
Z.~He, B.~Ai, Y.~Liu, W.~Wan, H.~I. Christensen, and H.~Su.
\newblock Learning dexterous deformable object manipulation through cross-embodiment dynamics learning.
\newblock In \emph{RSS Workshop on Dexterous Manipulation: Learning and Control with Diverse Data}, 2025.

\bibitem[Rong et~al.(2020)Rong, Bian, Xu, Xie, Wei, Huang, and Huang]{rong2020self}
Y.~Rong, Y.~Bian, T.~Xu, W.~Xie, Y.~Wei, W.~Huang, and J.~Huang.
\newblock Self-supervised graph transformer on large-scale molecular data.
\newblock \emph{Advances in neural information processing systems}, 33:\penalty0 12559--12571, 2020.

\bibitem[Ma et~al.(2024)Ma, Wang, Jia, Chen, Liu, Li, Chen, and Qiao]{ma2024latte}
X.~Ma, Y.~Wang, G.~Jia, X.~Chen, Z.~Liu, Y.-F. Li, C.~Chen, and Y.~Qiao.
\newblock Latte: Latent diffusion transformer for video generation.
\newblock \emph{arXiv preprint arXiv:2401.03048}, 2024.

\bibitem[Liu et~al.(2023)Liu, Wu, Van~Hoorick, Tokmakov, Zakharov, and Vondrick]{liu2023zero}
R.~Liu, R.~Wu, B.~Van~Hoorick, P.~Tokmakov, S.~Zakharov, and C.~Vondrick.
\newblock Zero-1-to-3: Zero-shot one image to 3d object.
\newblock In \emph{Proceedings of the IEEE/CVF international conference on computer vision}, pages 9298--9309, 2023.

\bibitem[R{\"u}hling~Cachay et~al.(2023)R{\"u}hling~Cachay, Zhao, Joren, and Yu]{ruhling2023dyffusion}
S.~R{\"u}hling~Cachay, B.~Zhao, H.~Joren, and R.~Yu.
\newblock Dyffusion: A dynamics-informed diffusion model for spatiotemporal forecasting.
\newblock \emph{Advances in neural information processing systems}, 36:\penalty0 45259--45287, 2023.

\bibitem[Chi et~al.(2023)Chi, Feng, Du, Xu, Cousineau, Burchfiel, and Song]{chi2023diffusionpolicy}
C.~Chi, S.~Feng, Y.~Du, Z.~Xu, E.~Cousineau, B.~Burchfiel, and S.~Song.
\newblock Diffusion policy: Visuomotor policy learning via action diffusion.
\newblock In \emph{Proceedings of Robotics: Science and Systems (RSS)}, 2023.

\bibitem[Vaswani et~al.(2017)Vaswani, Shazeer, Parmar, Uszkoreit, Jones, Gomez, Kaiser, and Polosukhin]{Vaswani2017attention}
A.~Vaswani, N.~Shazeer, N.~Parmar, J.~Uszkoreit, L.~Jones, A.~N. Gomez, L.~Kaiser, and I.~Polosukhin.
\newblock Attention is all you need.
\newblock In \emph{Proceedings of the 31st International Conference on Neural Information Processing Systems}, NIPS'17, page 6000–6010, Red Hook, NY, USA, 2017. Curran Associates Inc.
\newblock ISBN 9781510860964.

\bibitem[Matas et~al.(2018)Matas, James, and Davison]{pmlr-v87-matas18a}
J.~Matas, S.~James, and A.~J. Davison.
\newblock Sim-to-real reinforcement learning for deformable object manipulation.
\newblock In A.~Billard, A.~Dragan, J.~Peters, and J.~Morimoto, editors, \emph{Proceedings of The 2nd Conference on Robot Learning}, volume~87 of \emph{Proceedings of Machine Learning Research}, pages 734--743. PMLR, 29--31 Oct 2018.
\newblock URL \url{https://proceedings.mlr.press/v87/matas18a.html}.

\bibitem[Jangir et~al.(2020)Jangir, Alenyà, and Torras]{jan2020rl}
R.~Jangir, G.~Alenyà, and C.~Torras.
\newblock Dynamic cloth manipulation with deep reinforcement learning.
\newblock In \emph{2020 IEEE International Conference on Robotics and Automation (ICRA)}, pages 4630--4636, 2020.
\newblock \doi{10.1109/ICRA40945.2020.9196659}.

\bibitem[Avigal et~al.(2022)Avigal, Berscheid, Asfour, Kröger, and Goldberg]{9981402}
Y.~Avigal, L.~Berscheid, T.~Asfour, T.~Kröger, and K.~Goldberg.
\newblock Speedfolding: Learning efficient bimanual folding of garments.
\newblock In \emph{2022 IEEE/RSJ International Conference on Intelligent Robots and Systems (IROS)}, pages 1--8, 2022.
\newblock \doi{10.1109/IROS47612.2022.9981402}.

\bibitem[Fu et~al.(2024)Fu, Zhao, and Finn]{fu2024mobile}
Z.~Fu, T.~Z. Zhao, and C.~Finn.
\newblock Mobile aloha: Learning bimanual mobile manipulation with low-cost whole-body teleoperation.
\newblock In \emph{{Conference on Robot Learning (CoRL)}}, 2024.

\bibitem[Ze et~al.(2024)Ze, Zhang, Zhang, Hu, Wang, and Xu]{Ze2024DP3}
Y.~Ze, G.~Zhang, K.~Zhang, C.~Hu, M.~Wang, and H.~Xu.
\newblock 3d diffusion policy: Generalizable visuomotor policy learning via simple 3d representations.
\newblock In \emph{Proceedings of Robotics: Science and Systems (RSS)}, 2024.

\bibitem[Pertsch et~al.(2025)Pertsch, Stachowicz, Ichter, Driess, Nair, Vuong, Mees, Finn, and Levine]{pertsch2025fast}
K.~Pertsch, K.~Stachowicz, B.~Ichter, D.~Driess, S.~Nair, Q.~Vuong, O.~Mees, C.~Finn, and S.~Levine.
\newblock Fast: Efficient action tokenization for vision-language-action models.
\newblock 2025.
\newblock URL \url{https://doi.org/10.48550/arXiv.2501.09747}.

\bibitem[Chen et~al.(2021)Chen, Ma, Lu, and Hsu]{Chen2021AbInitio}
S.~Chen, X.~Ma, Y.~Lu, and D.~Hsu.
\newblock Ab initio particle-based object manipulation.
\newblock In D.~A. Shell, M.~Toussaint, and M.~A. Hsieh, editors, \emph{Robotics: Science and Systems XVII, Virtual Event, July 12-16, 2021}, 2021.
\newblock \doi{10.15607/RSS.2021.XVII.071}.
\newblock URL \url{https://doi.org/10.15607/RSS.2021.XVII.071}.

\bibitem[Shi et~al.(2022)]{shi2022robocraft}
H.~Shi et~al.
\newblock Robocraft: Learning to see, simulate, and shape elasto-plastic objects with graph networks.
\newblock In \emph{Proceedings of Robotics: Science and Systems (RSS)}, 2022.

\bibitem[Shi et~al.(2023)Shi, Xu, Clarke, Li, and Wu]{shi2023robocook}
H.~Shi, H.~Xu, S.~Clarke, Y.~Li, and J.~Wu.
\newblock Robocook: Long-horizon elasto-plastic object manipulation with diverse tools.
\newblock In J.~Tan, M.~Toussaint, and K.~Darvish, editors, \emph{Conference on Robot Learning, CoRL 2023, 6-9 November 2023, Atlanta, GA, {USA}}, volume 229 of \emph{Proceedings of Machine Learning Research}, pages 642--660. {PMLR}, 2023.
\newblock URL \url{https://proceedings.mlr.press/v229/shi23a.html}.

\bibitem[Ai et~al.(2024)Ai, Tian, Shi, Wang, Tan, Li, and Wu]{ai2024robopack}
B.~Ai, S.~Tian, H.~Shi, Y.~Wang, C.~Tan, Y.~Li, and J.~Wu.
\newblock Robopack: Learning tactile-informed dynamics models for dense packing.
\newblock \emph{Robotics: Science and Systems (RSS)}, 2024.
\newblock URL \url{https://arxiv.org/abs/2407.01418}.

\bibitem[Huang et~al.(2024)Huang, Wang, Li, Zhang, and Fei-Fei]{huang2024rekep}
W.~Huang, C.~Wang, Y.~Li, R.~Zhang, and L.~Fei-Fei.
\newblock Rekep: Spatio-temporal reasoning of relational keypoint constraints for robotic manipulation.
\newblock \emph{arXiv preprint arXiv:2409.01652}, 2024.

\bibitem[Longhini et~al.(2024)Longhini, Welle, Erickson, and Kragic]{longhini2024adafold}
A.~Longhini, M.~C. Welle, Z.~Erickson, and D.~Kragic.
\newblock Adafold: Adapting folding trajectories of cloths via feedback-loop manipulation.
\newblock \emph{IEEE Robotics and Automation Letters}, 2024.

\bibitem[Ai et~al.(2025)Ai, Tian, Shi, Wang, Pfaff, Tan, Christensen, Su, Wu, and Li]{ai2025review}
B.~Ai, S.~Tian, H.~Shi, Y.~Wang, T.~Pfaff, C.~Tan, H.~I. Christensen, H.~Su, J.~Wu, and Y.~Li.
\newblock A review of learning-based dynamics models for robotic manipulation.
\newblock \emph{Science Robotics}, 2025.

\bibitem[Hoque et~al.(2022)Hoque, Seita, Balakrishna, Ganapathi, Tanwani, Jamali, Yamane, Iba, and Goldberg]{10.1007/s10514-021-10001-0}
R.~Hoque, D.~Seita, A.~Balakrishna, A.~Ganapathi, A.~K. Tanwani, N.~Jamali, K.~Yamane, S.~Iba, and K.~Goldberg.
\newblock Visuospatial foresight for physical sequential fabric manipulation.
\newblock \emph{Auton. Robots}, 46\penalty0 (1):\penalty0 175–199, Jan. 2022.
\newblock ISSN 0929-5593.
\newblock \doi{10.1007/s10514-021-10001-0}.
\newblock URL \url{https://doi.org/10.1007/s10514-021-10001-0}.

\bibitem[Yan et~al.(2021)Yan, Vangipuram, Abbeel, and Pinto]{pmlr-v155-yan21a}
W.~Yan, A.~Vangipuram, P.~Abbeel, and L.~Pinto.
\newblock Learning predictive representations for deformable objects using contrastive estimation.
\newblock In J.~Kober, F.~Ramos, and C.~Tomlin, editors, \emph{Proceedings of the 2020 Conference on Robot Learning}, volume 155 of \emph{Proceedings of Machine Learning Research}, pages 564--574. PMLR, 16--18 Nov 2021.
\newblock URL \url{https://proceedings.mlr.press/v155/yan21a.html}.

\bibitem[Yang et~al.(2023)Yang, Du, Ghasemipour, Tompson, Schuurmans, and Abbeel]{yang2023learning}
M.~Yang, Y.~Du, K.~Ghasemipour, J.~Tompson, D.~Schuurmans, and P.~Abbeel.
\newblock Learning interactive real-world simulators.
\newblock \emph{arXiv preprint arXiv:2310.06114}, 2023.

\bibitem[Ho et~al.(2020)Ho, Jain, and Abbeel]{ho2020denoising}
J.~Ho, A.~Jain, and P.~Abbeel.
\newblock Denoising diffusion probabilistic models.
\newblock \emph{arXiv preprint arxiv:2006.11239}, 2020.

\bibitem[Rombach et~al.(2022)Rombach, Blattmann, Lorenz, Esser, and Ommer]{Rombach_2022_CVPR}
R.~Rombach, A.~Blattmann, D.~Lorenz, P.~Esser, and B.~Ommer.
\newblock High-resolution image synthesis with latent diffusion models.
\newblock In \emph{Proceedings of the IEEE/CVF Conference on Computer Vision and Pattern Recognition (CVPR)}, pages 10684--10695, June 2022.

\bibitem[Peebles and Xie(2022)]{Peebles2022DiT}
W.~Peebles and S.~Xie.
\newblock Scalable diffusion models with transformers.
\newblock \emph{arXiv preprint arXiv:2212.09748}, 2022.

\bibitem[Wang et~al.(2024)Wang, Chen, Ma, Zhou, Huang, Wang, Yang, He, Yu, Yang, et~al.]{wang2023lavie}
Y.~Wang, X.~Chen, X.~Ma, S.~Zhou, Z.~Huang, Y.~Wang, C.~Yang, Y.~He, J.~Yu, P.~Yang, et~al.
\newblock Lavie: High-quality video generation with cascaded latent diffusion models.
\newblock \emph{IJCV}, 2024.

\bibitem[Poole et~al.(2022)Poole, Jain, Barron, and Mildenhall]{poole2022dreamfusion}
B.~Poole, A.~Jain, J.~T. Barron, and B.~Mildenhall.
\newblock Dreamfusion: Text-to-3d using 2d diffusion.
\newblock \emph{arXiv}, 2022.

\bibitem[Alonso et~al.(2024)Alonso, Jelley, Micheli, Kanervisto, Storkey, Pearce, and Fleuret]{alonso2024diffusionworldmodelingvisual}
E.~Alonso, A.~Jelley, V.~Micheli, A.~Kanervisto, A.~Storkey, T.~Pearce, and F.~Fleuret.
\newblock Diffusion for world modeling: Visual details matter in atari.
\newblock In \emph{Thirty-eighth Conference on Neural Information Processing Systems}, 2024.
\newblock URL \url{https://arxiv.org/abs/2405.12399}.

\bibitem[Ding et~al.(2024)Ding, Zhang, Tian, and Zheng]{ding2024diffusion}
Z.~Ding, A.~Zhang, Y.~Tian, and Q.~Zheng.
\newblock Diffusion world model: Future modeling beyond step-by-step rollout for offline reinforcement learning.
\newblock \emph{arXiv preprint arXiv:2402.03570}, 2024.

\bibitem[Chen et~al.(2021)Chen, Sax, Lewis, Armeni, Savarese, Zamir, Malik, and Pinto]{pmlr-v155-chen21f}
B.~Chen, A.~Sax, F.~Lewis, I.~Armeni, S.~Savarese, A.~Zamir, J.~Malik, and L.~Pinto.
\newblock Robust policies via mid-level visual representations: An experimental study in manipulation and navigation.
\newblock In J.~Kober, F.~Ramos, and C.~Tomlin, editors, \emph{Proceedings of the 2020 Conference on Robot Learning}, volume 155 of \emph{Proceedings of Machine Learning Research}, pages 2328--2346. PMLR, 16--18 Nov 2021.
\newblock URL \url{https://proceedings.mlr.press/v155/chen21f.html}.

\bibitem[Ai et~al.(2023)Ai, Wu, and Hsu]{ai2023invariance}
B.~Ai, Z.~Wu, and D.~Hsu.
\newblock Invariance is key to generalization: Examining the role of representation in sim-to-real transfer for visual navigation.
\newblock In \emph{International Symposium on Experimental Robotics}, pages 69--80. Springer, 2023.

\bibitem[Qi et~al.(2016)Qi, Su, Mo, and Guibas]{qi2016pointnet}
C.~R. Qi, H.~Su, K.~Mo, and L.~J. Guibas.
\newblock Pointnet: Deep learning on point sets for 3d classification and segmentation.
\newblock \emph{arXiv preprint arXiv:1612.00593}, 2016.

\bibitem[Dosovitskiy et~al.(2021)Dosovitskiy, Beyer, Kolesnikov, Weissenborn, Zhai, Unterthiner, Dehghani, Minderer, Heigold, Gelly, Uszkoreit, and Houlsby]{dosovitskiy2020vit}
A.~Dosovitskiy, L.~Beyer, A.~Kolesnikov, D.~Weissenborn, X.~Zhai, T.~Unterthiner, M.~Dehghani, M.~Minderer, G.~Heigold, S.~Gelly, J.~Uszkoreit, and N.~Houlsby.
\newblock An image is worth 16x16 words: Transformers for image recognition at scale.
\newblock \emph{ICLR}, 2021.

\bibitem[Xu et~al.(2019)Xu, Sun, Zhang, Zhao, and Lin]{xu2019understanding}
J.~Xu, X.~Sun, Z.~Zhang, G.~Zhao, and J.~Lin.
\newblock Understanding and improving layer normalization.
\newblock \emph{Advances in neural information processing systems}, 32, 2019.

\bibitem[Mildenhall et~al.(2021)Mildenhall, Srinivasan, Tancik, Barron, Ramamoorthi, and Ng]{mildenhall2021nerf}
B.~Mildenhall, P.~P. Srinivasan, M.~Tancik, J.~T. Barron, R.~Ramamoorthi, and R.~Ng.
\newblock Nerf: Representing scenes as neural radiance fields for view synthesis.
\newblock \emph{Communications of the ACM}, 65\penalty0 (1):\penalty0 99--106, 2021.

\bibitem[Williams et~al.(2016)Williams, Drews, Goldfain, Rehg, and Theodorou]{7487277}
G.~Williams, P.~Drews, B.~Goldfain, J.~M. Rehg, and E.~A. Theodorou.
\newblock Aggressive driving with model predictive path integral control.
\newblock In \emph{2016 IEEE International Conference on Robotics and Automation (ICRA)}, pages 1433--1440, 2016.
\newblock \doi{10.1109/ICRA.2016.7487277}.

\bibitem[Xiang et~al.(2020)Xiang, Qin, Mo, Xia, Zhu, Liu, Liu, Jiang, Yuan, Wang, Yi, Chang, Guibas, and Su]{SAPIEN}
F.~Xiang, Y.~Qin, K.~Mo, Y.~Xia, H.~Zhu, F.~Liu, M.~Liu, H.~Jiang, Y.~Yuan, H.~Wang, L.~Yi, A.~X. Chang, L.~J. Guibas, and H.~Su.
\newblock {SAPIEN}: A simulated part-based interactive environment.
\newblock In \emph{The IEEE Conference on Computer Vision and Pattern Recognition (CVPR)}, June 2020.

\bibitem[Bouaziz et~al.(2014)Bouaziz, Martin, Liu, Kavan, and Pauly]{Bouaziz2014ProjectiveDynamics}
S.~Bouaziz, S.~Martin, T.~Liu, L.~Kavan, and M.~Pauly.
\newblock Projective dynamics: fusing constraint projections for fast simulation.
\newblock \emph{ACM Trans. Graph.}, 33\penalty0 (4), July 2014.
\newblock ISSN 0730-0301.
\newblock \doi{10.1145/2601097.2601116}.
\newblock URL \url{https://doi.org/10.1145/2601097.2601116}.

\bibitem[Ly et~al.(2020)Ly, Jouve, Boissieux, and Bertails-Descoubes]{Ly2020DryFrictionalPD}
M.~Ly, J.~Jouve, L.~Boissieux, and F.~Bertails-Descoubes.
\newblock Projective dynamics with dry frictional contact.
\newblock \emph{ACM Trans. Graph.}, 39\penalty0 (4), Aug. 2020.
\newblock ISSN 0730-0301.
\newblock \doi{10.1145/3386569.3392396}.
\newblock URL \url{https://doi.org/10.1145/3386569.3392396}.

\bibitem[Zhang et~al.(2023)Zhang, Chen, Li, Xiang, Qin, Gu, Ling, Liu, Zeng, Han, Huang, Mu, Xu, and Su]{10027470}
X.~Zhang, R.~Chen, A.~Li, F.~Xiang, Y.~Qin, J.~Gu, Z.~Ling, M.~Liu, P.~Zeng, S.~Han, Z.~Huang, T.~Mu, J.~Xu, and H.~Su.
\newblock Close the optical sensing domain gap by physics-grounded active stereo sensor simulation.
\newblock \emph{IEEE Transactions on Robotics}, pages 1--19, 2023.
\newblock \doi{10.1109/TRO.2023.3235591}.

\bibitem[Bertiche et~al.(2020)Bertiche, Madadi, and Escalera]{bertiche2020cloth3d}
H.~Bertiche, M.~Madadi, and S.~Escalera.
\newblock Cloth3d: clothed 3d humans.
\newblock In \emph{European Conference on Computer Vision}, pages 344--359. Springer, 2020.

\bibitem[Minderer et~al.(2024)Minderer, Gritsenko, and Houlsby]{minderer2024scaling}
M.~Minderer, A.~Gritsenko, and N.~Houlsby.
\newblock Scaling open-vocabulary object detection.
\newblock \emph{Advances in Neural Information Processing Systems}, 36, 2024.

\bibitem[Kirillov et~al.(2023)Kirillov, Mintun, Ravi, Mao, Rolland, Gustafson, Xiao, Whitehead, Berg, Lo, Doll{\'a}r, and Girshick]{kirillov2023segany}
A.~Kirillov, E.~Mintun, N.~Ravi, H.~Mao, C.~Rolland, L.~Gustafson, T.~Xiao, S.~Whitehead, A.~C. Berg, W.-Y. Lo, P.~Doll{\'a}r, and R.~Girshick.
\newblock Segment anything.
\newblock \emph{arXiv:2304.02643}, 2023.

\end{thebibliography}
